\documentclass[journal]{vgtc}                          

\usepackage{color}

\ifpdf
  \pdfoutput=1\relax                   
  \usepackage{graphicx}                
  \DeclareGraphicsExtensions{.pdf,.png,.jpg,.jpeg} 
\else
  \usepackage{graphicx}                
  \DeclareGraphicsExtensions{.eps}     
\fi%

\graphicspath{{figures/}{pictures/}{images/}{./}} 

\usepackage{microtype}                 
\PassOptionsToPackage{warn}{textcomp}  
\usepackage{textcomp}                  
\usepackage{mathptmx}                  
\usepackage{times}                     
\usepackage{cite}                      
\usepackage{tabu}                      
\usepackage{booktabs}                  

\usepackage[utf8]{inputenc}
\usepackage{authblk}
\usepackage{amsmath}
\usepackage{multirow}
\usepackage{float}

\usepackage[switch]{lineno}

\onlineid{206}

\title{Deep 6-DOF Tracking}

\author{Mathieu Garon and Jean-François Lalonde}
\authorfooter{

\item Mathieu Garon is with Université Laval. E-mail: mathieu.garon.2@ulaval.ca.
\item Jean-François Lalonde is with Université Laval. E-mail: jflalonde@gel.ulaval.ca.
}

\shortauthortitle{Garon and Lalonde: Deep 6-DOF Tracking}

\abstract{We present a temporal 6-DOF tracking method which leverages deep learning to achieve state-of-the-art performance on challenging datasets of real world capture. Our method is both more accurate and more robust to occlusions than the existing best performing approaches while maintaining real-time performance. To assess its efficacy, we evaluate our approach on several challenging RGBD sequences of real objects in a variety of conditions. Notably, we systematically evaluate robustness to occlusions through a series of sequences where the object to be tracked is increasingly occluded. Finally, our approach is purely data-driven and does not require any hand-designed features: robust tracking is automatically learned from data. } 


\teaser{
\centering
\includegraphics[width=\linewidth]{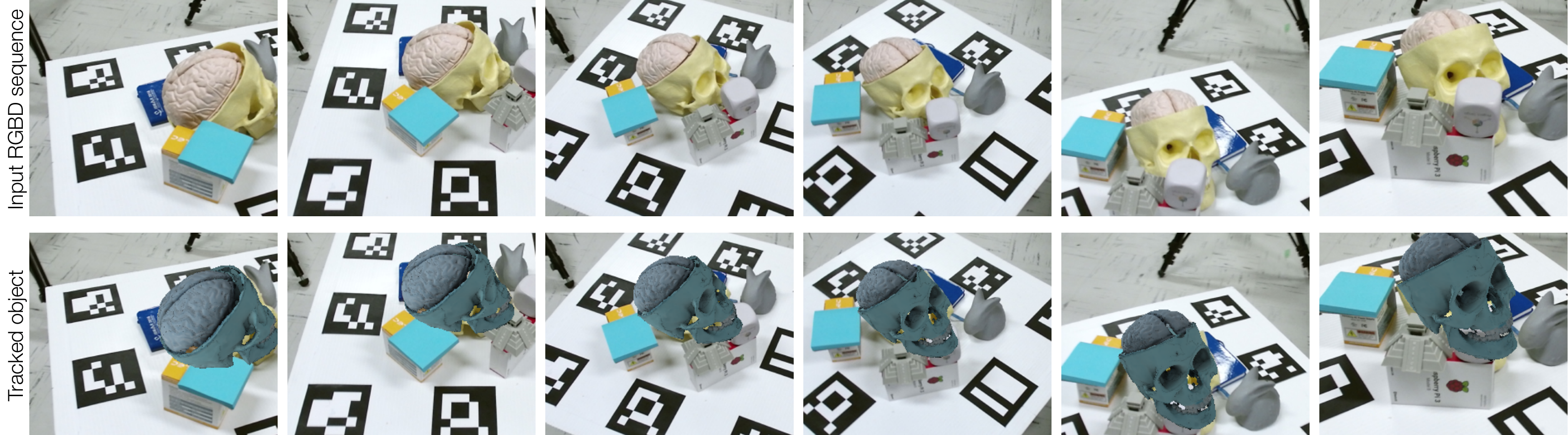}
\caption{From an input RGBD sequence (top), our method leverages a deep neural network to automatically track the 6-DOF pose of an object even under significant clutter and occlusion (bottom). We demonstrate, through extensive experiments on a novel dataset of real objects with known ground truth pose, that our approach outperforms the state of the art both in terms of accuracy and robustness to occlusions.}
\label{fig:teaser}
}

\vgtcinsertpkg

\begin{document}

\maketitle
\section{Introduction}

The\let\thefootnote\relax\footnote{TVCG preprint (DOI:TVCG2734599) © 2017 IEEE. Personal use of this material is permitted. Permission from IEEE must be obtained for all other uses, in any current or future media, including reprinting/republishing this material for advertising or promotional purposes, creating new collective works, for resale or redistribution to servers or lists, or reuse of any copyrighted component of this work in other works} recent advent of 3D-enabled AR devices is enabling a whole range of new applications. In addition to helping in robustly positioning the camera in the environment via SLAM-based techniques, RGBD sensors also allow the system to track objects moving freely in space by providing estimates of their 6-DOF poses at each frame. This is a well-known, but notably hard problem. Indeed, a successful 6-DOF object tracker must be accurate, stable, achieve real-time performance, and be robust to perturbations such as noise and occlusions. 

A successful series of approaches recently introduced by Tan et al.~\cite{tan2014multi,tan-iccv-15} propose to frame the RGBD tracking problem as one of learning, where the task is to learn the relationship between a pair of input frames and the relative 6-DOF pose change of the object in both of these frames. This ``temporal tracking by learning'' approach was demonstrated to be successful at achieving robust, real-time tracking results on real-world objects. However, we make the observation that while these approaches are suitable for sequences with relatively small levels of occlusions, they fail for larger levels of occlusion. Notably, our experiments demonstrate that hiding 20\% or more of the object often results in catastrophic failure, from which the tracker never recovers. For robust AR applications to work, we need a real-time tracker which is more robust to occlusions, and which will not generate these irrecoverable situations.

In this work, we present an accurate, real-time temporal 6-DOF object tracking method which is more robust to occlusions than existing state-of-the-art algorithms. Of particular interest, when occlusion is severe, our approach is very robust in estimating the \emph{position} of the object even when the rotation components fail. Thus, catastrophic failures occur much less frequently than existing approaches. \autoref{fig:teaser} shows qualitative results on a sequence with such occlusions.

Our main key contribution is to frame 6-DOF tracking as a deep learning problem. This contribution provides us with three key benefits. First, deep learning architectures can be trained on very large amounts of data, so they can be robust to a wide variety of capture conditions such as color shifts, illumination changes, motion blur, and occlusions. Second, they possess very efficient GPU implementations that can be processed in real-time on mobile GPUs given a small enough network. Finally, and perhaps most importantly, no hand-designed features need to be computed: object-specific features can automatically be learned from data. This is in contrast to most previous work (e.g. Tan et al.~\cite{tan2014multi,tan-iccv-15}) which compute specific, hand-designed features. 

Applying a deep convolutional neural network (CNN) to tracking is not trivial. Indeed, temporal tracking differs from ``tracking by detection'' in that the temporal tracker uses two frames adjacent in time, and assumes knowledge of the object pose at the previous frame. To train a deep network on that task, one could straightforwardly use the current and previous RGBD frames directly as input. Unfortunately, while doing so yields low prediction errors on a ``conventional'' machine learning test set (composed of pairs of frames as input and rigid pose change as target), it completely fails to track in sequences of multiple frames. Indeed, since the network never learned to correct itself, small errors accumulate rapidly and tracking is lost in a matter of milliseconds. Another solution could be to provide the previous estimate of the pose change instead of the previous RGBD frame as input to the network. In this case, this information alone is not rich enough to enable the network to learn robust high level representations, also yielding high tracking errors. To solve this problem, we propose to use an \emph{estimate} of the object pose from the previous timestep in the sequence as input to the network, in addition to the current RGBD frame. This allows the network to correct errors made in closed loop tracking. The feedback, which is the estimate of the current object pose, is obtained by rendering a synthetic RGBD frame of the tracked object. Thus, our approach requires a 3D model of the object a priori, and the tracker is trained for a specific object.

We make the following key contributions. First, we show a practical way of framing the 6-DOF tracking from RGBD data in a way that is suitable for training with deep networks. Second, we present a real-time tracking approach that is both more accurate, stable and more robust to occlusions and fast camera motion than the state-of-the-art. Third, we introduce a dataset\footnote{See {\scriptsize \url{http://www.jflalonde.ca/projects/deepTracking}}.} of calibrated real sequences of four objects captured by a Kinect v2 under different conditions. In particular, our dataset includes sequences in which the occlusion is slowly varied from low to high levels of occlusions (10\%--40\% of the object occluded). Since the only available tracking dataset~\cite{akkaladevi2016tracking} provides only an evaluation for fast moving cameras, we hope that a dataset benchmarking robustness to occlusions will spur research in this area. Finally, we provide an extensive comparison of our work with that of Tan et al.~\cite{tan-iccv-15} and Akkaladevi et al.~\cite{akkaladevi2016tracking}, which demonstrates the wide applicability of our approach. 

\section{Related Work}

\begin{figure*}[!h]
\centering
\footnotesize
\includegraphics[width=\linewidth]{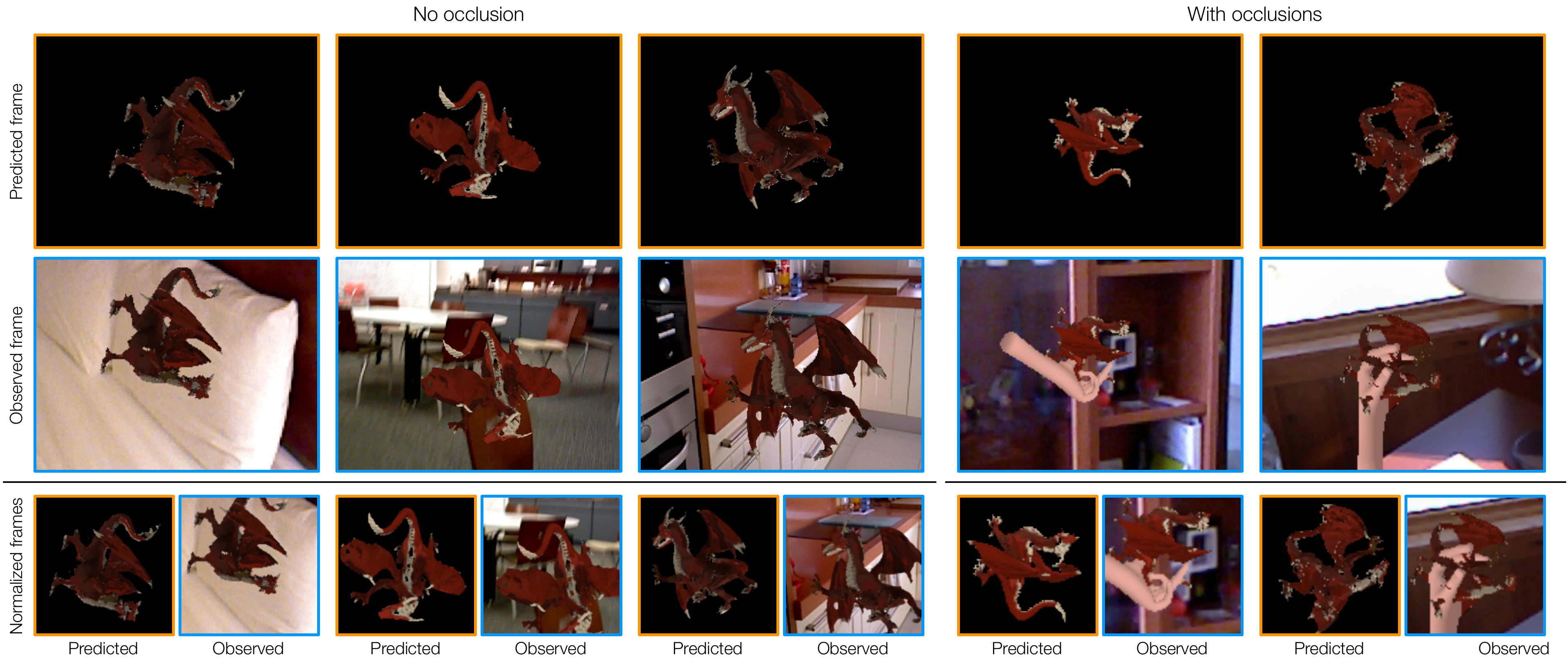}
\caption[]{Examples of images generated to train the CNN to track the dragon model. Examples without (leftmost three examples) and with (rightmost two examples) occlusions are illustrated. The top row shows the predicted frame $\mathbf{x}_\textrm{pred}$, and the second row the observed frame $\mathbf{x}_\textrm{obs}$, obtained by compositing a rendering of the object onto a real background from the SUN3D database~\cite{xiao2013sun3d}. The bottom row shows the normalized frames that are given as input to the CNN. }
\label{fig:data-generation}
\end{figure*}

Object tracking has received extensive attention in the literature. Here, we restrict our discussion to 6-DOF tracking (typically via RGBD data), and on deep learning applied to 2D tracking and 3D pose estimation. 

Tracking objects with 6-DOF is a highly geometric problem that can be solved with techniques such as ICP~\cite{newcombe2011kinectfusion, aldoma2013multimodal}. However, tracking smaller objects with higher accuracy requires more sophisticated methods such as particle filters with handcrafted features as likelihood estimation~\cite{kwon2007particle, Choi2013}. Recent methods leverage data-driven methods such as Random Forests~\cite{breiman2001random} to learn more robust features that better handle occlusion, and with low computing overhead. Of note, Krull et al.~\cite{Krull2014,krull2015learning,krull2016poseagent} use a more sophisticated likelihood function by regressing a pixel wise probability of the object and its local object coordinates. They use this representation in different frameworks such as particle filters, RANSAC and deep learning. Tan et al.~\cite{tan2014multi,tan-iccv-15} propose a complete learning-based method that uses very low CPU overhead. They combine multiple decision trees that regress a random subset of 3D points displacement in the object coordinate system, thus using only depth data for tracking. To limit the impact of occlusions, they propose to randomly select points near random edges of the object, hoping that occluding a part of the object will not affect those points, however it fails when the occluder does not follow this rule. This approach has later been expanded by Akkaladevi et al.~\cite{akkaladevi2016tracking}. While they use a similar idea to track the objects, they also propose a detection method to reinitialize the tracker at runtime. They provide a challenging dataset of a single sequence of a Primesense camera rapidly moving in front of 4 static objects on which we compare our approach (~\autoref{sec:dataset-of-akkaladevi}). Similarly to these works, we also employ a full data-driven solution. However, our deep CNN is much more robust to occlusions and naturally exploits multiple modalities such as RGBD, while still allowing real-time performance. Finally, access to IMU sensor data on mobile devices enabled better algorithms for camera tracking~\cite{tanskanen2013live,brunetto2015fusion}. Even though this data can help stabilize tracking, our method cannot directly benefit from inertial sensors since we track the \emph{relative} pose between the camera and the object, and not the camera pose only.

Previously, convolutional neural networks have been used for 2D object tracking via temporal methods~\cite{gan2015first, bertinetto2016fully} and real-time ``tracking by detection'' methods as well~\cite{redmon2016you, ren2015faster} on mobile GPUs. However, these methods focus on tracking the 2D bounding box in the image plane, whereas we track the 6-DOF pose of the object. Deep neural networks have also been used to solve 3D geometric problems such as estimating the pose of a known object in a single image~\cite{wohlhart2015learning, doumanoglou2016siamese,brachmann2016uncertainty,pavlakos20176}, learning robust descriptors for pose estimation~\cite{kehl2016deep,crivellaro2015novel,krull2015learning}, inferring a transformation between two input images~\cite{oberweger2015training,detone2016deep}, or estimating the camera pose from a single RGB image~\cite{kendall-iccv-15}. Oberweger et al.~\cite{oberweger2015training} proposed a feedback loop using a generator network with similar ideas to those described in this paper. While they propose to train a generator with a deep neural network architecture, we use a geometric rendering pipeline that generates stable and controlled input samples. While most of these methods split the descriptor learning and test phases, we provide an end-to-end method that fully leverages the representational power of deep architectures. To the best of our knowledge, we are the first to use deep learning for 6-DOF temporal object tracking.
\section{Training data generation}
\label{sec:data-generation}

As in~\cite{tan-iccv-15}, we employ a rendering-based method to generate the data necessary to train our deep network. Our network takes in two inputs: 1) a frame showing the object at the \emph{predicted} position, based on the estimation obtained from the previous timestamp; and 2) a frame of the object at the actual position, as \emph{observed} by a camera. To encourage the network to be robust to a variety of situations, we synthesize both these frames by rendering a 3D model of the object and simulating realistic capture conditions including object positions, backgrounds, noise, and lighting. Note that more accurate geometry and texture model will render samples that look more realistic, thus improving the overall performance of the network. In our experiments, we show that our approach successfully works with 3D models made with a Primesense sensor (from the PROFACTOR 3D dataset~\cite{akkaladevi2016tracking}) and with much more precise Creaform GoScan\textsuperscript{TM} sensors (in our dataset).

\subsection{Sampling random camera poses}
\label{sec:sampling-camera}

First, the object is placed at the origin of the world reference frame, and a random camera pose in spherical coordinates $(\theta,\phi)$ is sampled: $\theta \sim \text{U}(-180^\circ, 180^\circ)$ and $\phi = \cos^{-1}(2x - 1)$, where $x \sim \text{U}(0, 1)$. Here, $\text{U}(a, b)$ indicates a uniform distribution in the $[a, b]$ interval. A random roll angle is also sampled uniformly $\gamma \sim \text{U}(-180^\circ, 180^\circ)$. The camera is then displaced along the ray between the camera pose and the origin by a random amount $r \sim \text{U}(0.4\text{m}, 1.5\text{m})$ to obtain the \emph{observed} pose $\mathbf{p}_\text{obs}$.

The \emph{predicted} camera pose $\mathbf{p}_\text{pred}$ is obtained by applying a random, 6-DOF rigid transformation from $\mathbf{p}_\text{obs}$. This is obtained by sampling a random translation $t_{x,y,z} \sim \text{U}(-20\text{mm}, 20\text{mm})$ and rotation $r_{\alpha,\beta,\gamma} \sim \text{U}(-10^\circ, 10^\circ)$ in Euler angle notation. The \emph{inverse} of this random transformation is applied to $\mathbf{p}_\text{obs}$ to obtain $\mathbf{p}_\text{pred}$. Thus, the target label (displacement between the two poses) is the 6-vector $\mathbf{y} = [t_x \;\; t_y \;\; t_z \;\; r_\alpha \;\; r_\beta \;\; r_\gamma]$ of concatenated translations and Euler angles representing the object transformation in the camera reference frame. 

\subsection{Rendering pipeline}

The proposed rendering pipeline takes in a textured 3D model of the object to track, and the two previously-defined camera poses $\mathbf{p}_\text{pred}$ and $\mathbf{p}_\text{obs}$. Examples of training images obtained with this technique are shown in~\autoref{fig:data-generation}. 

The predicted image $\mathbf{x}_\text{pred}$ is obtained by rendering the object on its own by placing the virtual camera at $\mathbf{p}_\text{pred}$. The object is rendered using ambient occlusions, and lit with a combination of an ambient and a directional white light source of intensity 0.65 and 0.4 respectively. The directional light source points downwards with respect to the camera viewing direction. 

The observed image $\mathbf{x}_\text{obs}$ is obtained by rendering the object by placing the virtual camera at $\mathbf{p}_\text{obs}$. Here, the light source direction is sampled uniformly on the sphere (using the same process as in~\autoref{sec:sampling-camera}). To simulate more realistic capture conditions, the resulting image is then composited onto an RGBD background. The background is obtained by randomly selecting a frame in the SUN3D dataset~\cite{xiao2013sun3d}, which is a dataset of 415 RGBD sequences captured in 254 different places. Since this is a dataset of video sequences, many frames are extremely similar to one another. To circumvent this issue, we greedily select a set of different frames by starting with a reference image and by computing the SSIM image similarity metric~\cite{wang-tip-04} between the subsequent frames in the video. A frame is selected as ``different'' if the SSIM is above a certain threshold. The process is then repeated by selecting this frame as the reference. 

\subsection{Data augmentation}

Additional perturbations are applied during training to increase the robustness of the CNN, namely: color, noise, blur, and occluders. Every augmentation is applied randomly online after loading the minibatch to enrich the training data distribution.

To simulate small color shifts between the render and sensor, a random perturbation of $h \sim \text{U}(-.05, .05)$ is applied to the hue and luminosity (in the HSV color space) of the object. The noise is generated with a gaussian distribution $\text{N}(0, \sigma)$, where $\sigma \sim \text{U}(0, 2)$, and is added to all of the RGBD channels of the image. This process is applied to a random subset of 95\% of the training images (the other 5\% remain noiseless). We found that this method increased the tolerance of the network to noise, while preventing overfitting (which occured if the noise properties were the same on all training images). To account for rapid camera movement, 40\% of the training images are blurred with a $3 \times 3$ mean filter. The kernel is applied to all RGBD channels. 

To increase robustness to occlusion, the training data is augmented by occluding the object with another virtual object. To do so, a random rigid transformation is sampled, applied to the object pose, and used to render a 3D model. For the experimentation a hand model is used with its color set by randomly sampling a hue and a luminosity in the HSV color space, and converting back to RGB. The model is then composited onto the render, by using the depth channel as z-buffer. This procedure is applied randomly on 60\% of the training data. The last two columns of~\autoref{fig:data-generation} show some example training data with a randomly generated occluder. 

\subsection{Data normalization}

The neural network expects fixed size inputs of $150 \times 150$ pixels resolution (see~\autoref{sec:learning}). However, the object may have varying dimensions on the image plane as it moves around in 3D (see~\autoref{fig:data-generation}). It is possible to use the previous pose of the object to normalize the representation of the object on the image plane with the following method. 

A square bounding box is drawn around the object, and the resulting RGBD image is resized to the target resolution using bilinear interpolation. To account for large variations in pose changes, the bounding box is 15\% larger than the object size. This method brings the object to the same projection plane regardless of its pose w.r.t the camera. Once the projection is normalized for each channel, the depth pixels are shifted to the center of the object to ensure invariance to the object absolute depth.

All channels are also normalized individually by subtracting the mean and dividing by the standard deviation of a subset of the training dataset. Input frames normalized with this approach are illustrated in the bottom row of~\autoref{fig:data-generation}. Finally, the labels are scaled to the $[-1, 1]$ interval based on the range defined in~\autoref{sec:sampling-camera}.

\section{Learning}
\label{sec:learning}

Our approach relies on a convolutional neural network that learns to regress the relative 6-DOF pose of an object given two input RGBD images: 1) an image of the object at the predicted position (from the previous timestamp in the video sequence) $\mathbf{x}_\text{pred}$; and 2) an image of the observed object at the current timestamp $\mathbf{x}_\text{obs}$. It is trained on a rendered dataset of the current object, as described in~\autoref{sec:data-generation}.

\subsection{Network architecture}

\begin{figure}[!ht]
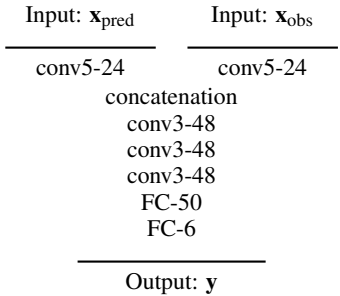

\centering
\begin{tabular}{cc}
Input: $\mathbf{x}_\text{pred}$ & Input: $\mathbf{x}_\text{obs}$ \\
\noindent\rule{2cm}{.8pt} & \noindent\rule{2cm}{.8pt} \\
conv5-24 & conv5-24
\end{tabular}
\\
\begin{tabular}{c}
concatenation \\
conv3-48 \\
conv3-48 \\
conv3-48 \\
FC-50    \\
FC-6	 \\
\noindent\rule{2.5cm}{.8pt} \\
Output: $\mathbf{y}$ \\
\end{tabular}
\caption[]{The architecture of our deep 6-DOF object tracker. Here, the notation ``conv$x$-$y$'' indicates a convolution layer of $y$ filters of dimension $x \times x$, and ``FC-$x$'' means a fully-connected layer of $x$ units. The network first convolves the predicted and observed frames $\mathbf{x}_\textrm{pred}$ and $\mathbf{x}_\textrm{obs}$ independently through a single convolution layer (with different weights). The output of these convolutions are then concatenated, and passed through three additional convolution layers. Finally, two fully-connected layers provide the estimated transformation between both frames $\mathbf{y}$ as output. Each convolution layer is followed by a max pooling $2 \times 2$ operation to downsample their representation. All layers (except the last FC-6) have batch normalization and the ELU activation function~\cite{clevert-iclr-16}. The activation function for that last layer is tanh. We use a dropout of 50\% on the input connections to the FC-50 layer.  }
\label{fig:architecture}
\end{figure}

The architecture of the proposed CNN is presented in concise form in~\autoref{fig:architecture}. Each input $\mathbf{x}_\text{pred}$ and $\mathbf{x}_\text{obs}$ is expected to be of $150 \times 150 \times 4$ resolution. They are first convoluted independently via a single convolution layer. The output of these layers is concatenated into a single feature map, which is subsequently fed to three convolution layers. Two fully-connected layers follow to produce the output $\mathbf{y}$, that is, the relative 6-DOF pose of the object between the two input frames.

\subsection{Training details}
\label{sec:training-details}

The network from~\autoref{fig:architecture} is trained by minimizing the mean squared error (MSE) loss between the prediction $\mathbf{y}$ and the ground truth target $\mathbf{t}$. For a given object, we use $75\%$ of the samples as training data, and $25\%$ as validation data. The validation data is used for early stopping, which acts as a form of regularization. 

We use the ADAM optimizer~\cite{kingma-iclr-15} with a minibatch size of $64$, initial learning rate of $0.005$, and learning rate decay of 1E-5. We first train the network with a synthetic dataset of 250,000 sample images generated with the approach described in~{\autoref{sec:data-generation}. Before testing on real data, we further fine-tune the weights on 180 real data samples (see~\autoref{sec:real-data-acquisition} for a description of how this real training data is acquired). The entire training time is 5 hours for initial training, and 30 minutes for fine-tuning on an Nvidia Tesla K40c GPU.

\section{Experiments}
\label{sec:experiments}

We now proceed to describe a series of experiments conducted to evaluate the accuracy and robustness of our approach on two challenging datasets of RGBD sequences. First, we compare the performance of our tracker on the existing PROFACTOR 3D dataset~\cite{akkaladevi2016tracking}. Then, we describe the data acquisition process for our new dataset, followed by a quantitative comparison on several different real objects. A systematic evaluation of robustness to occlusions and initialization follows, and we conclude with a discussion on speed and memory usage.

\begin{figure}[!t]
\centering
\footnotesize
\setlength{\tabcolsep}{1pt}
\begin{tabular}{cc}
 \includegraphics[width=.48\linewidth]{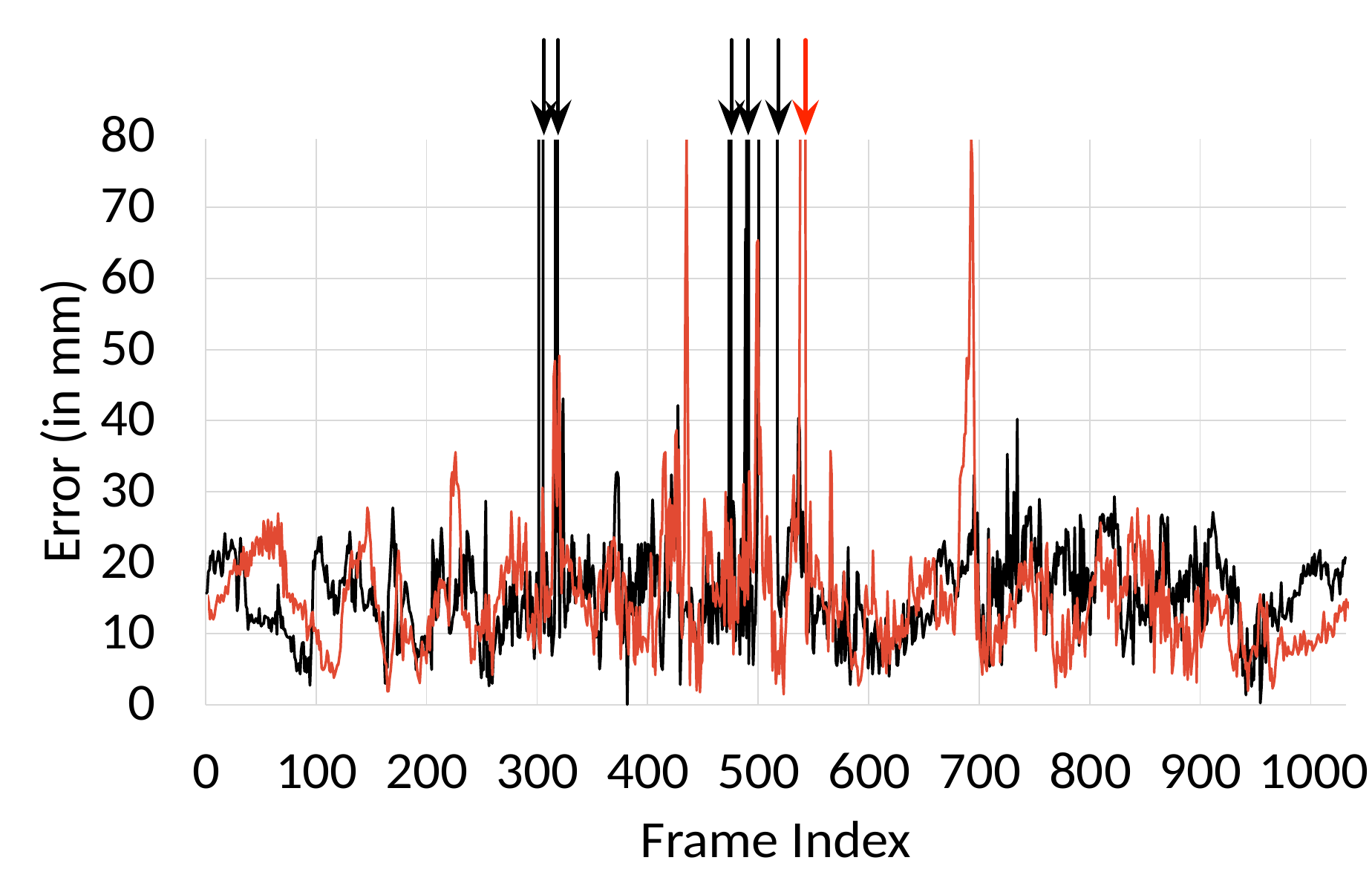} &
 \includegraphics[width=.48\linewidth]{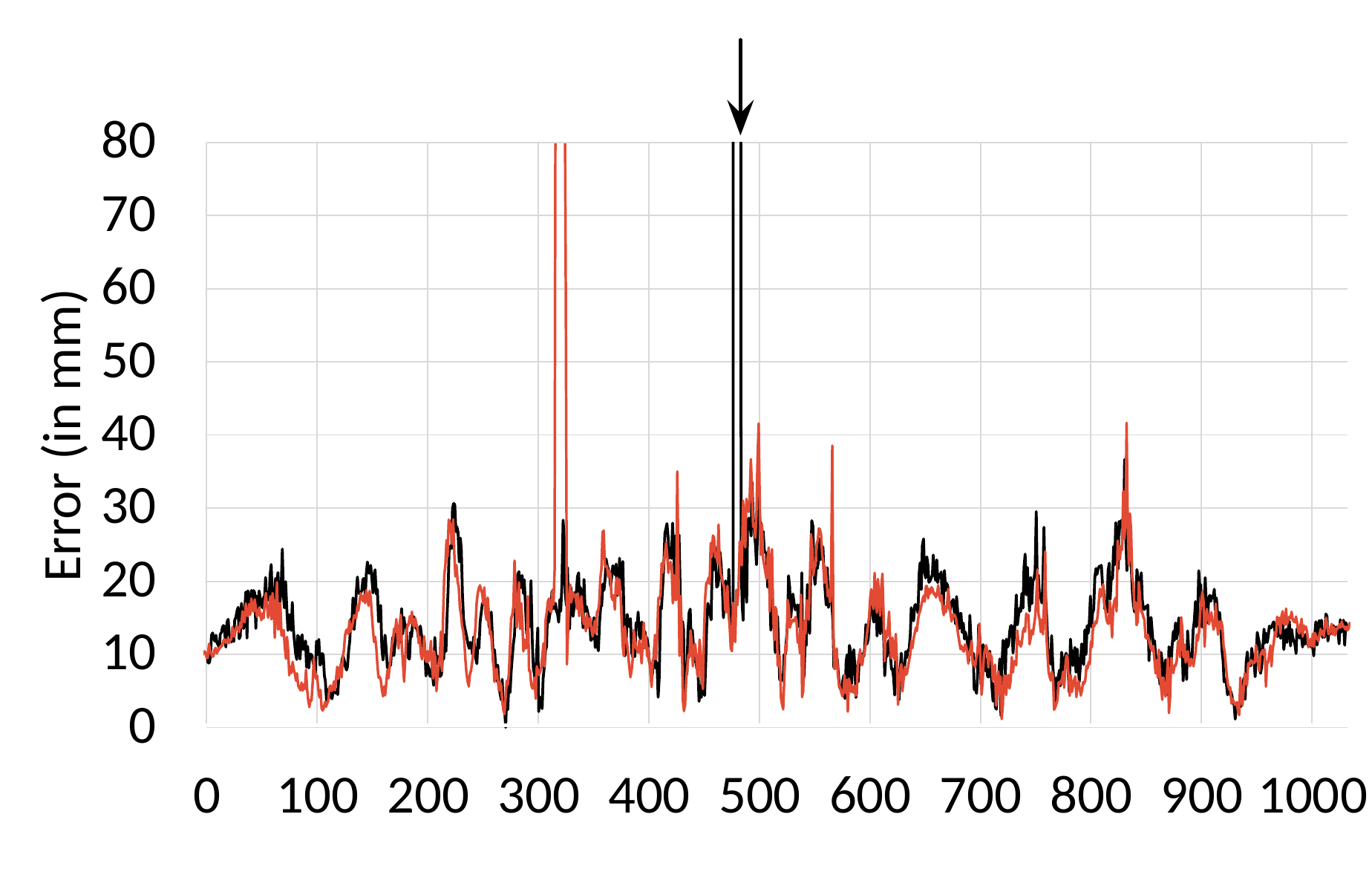} \\
(a) Bin & (b) Garden gnome \\
 \includegraphics[width=.48\linewidth]{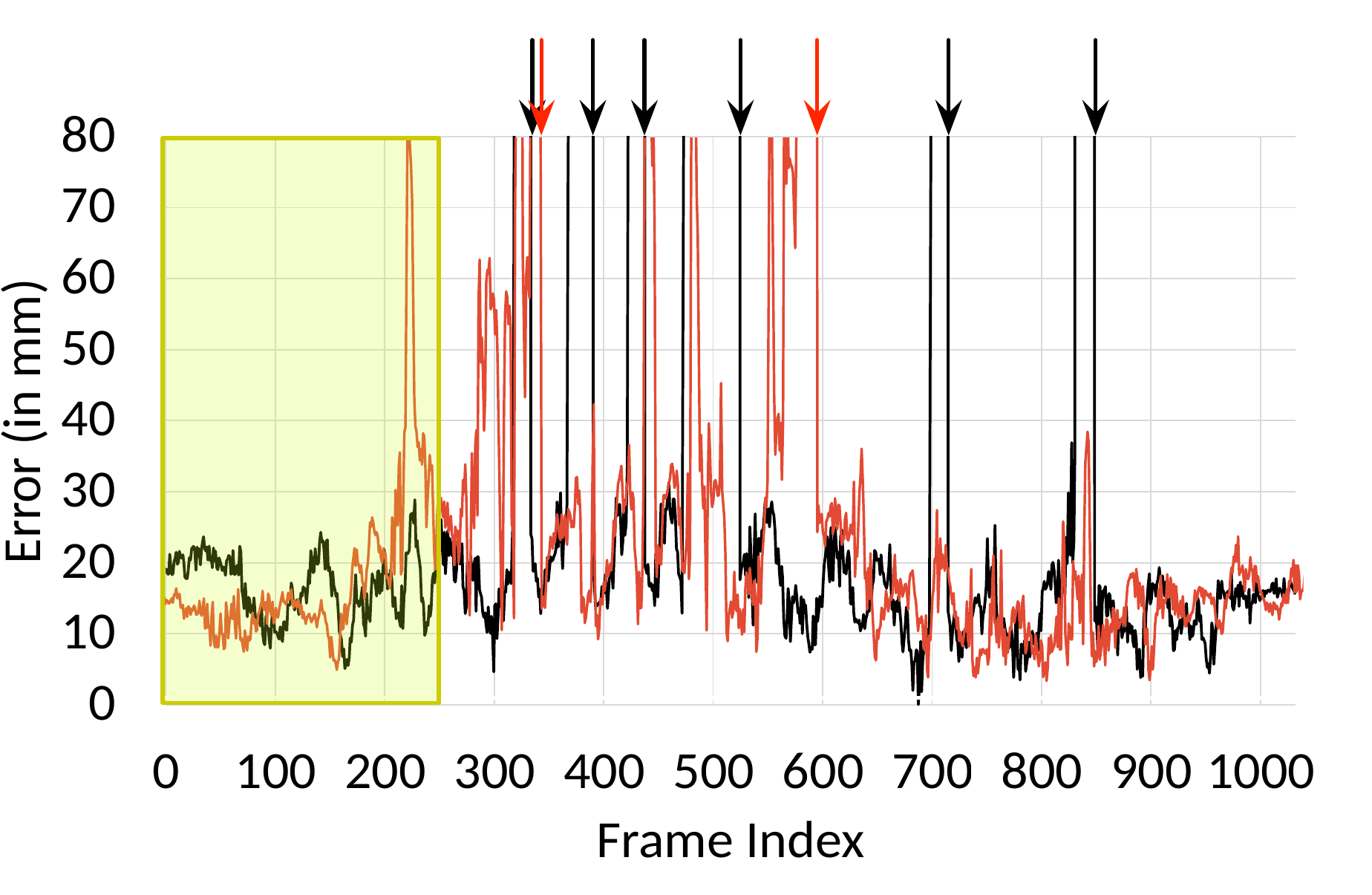} &
 \includegraphics[width=.48\linewidth]{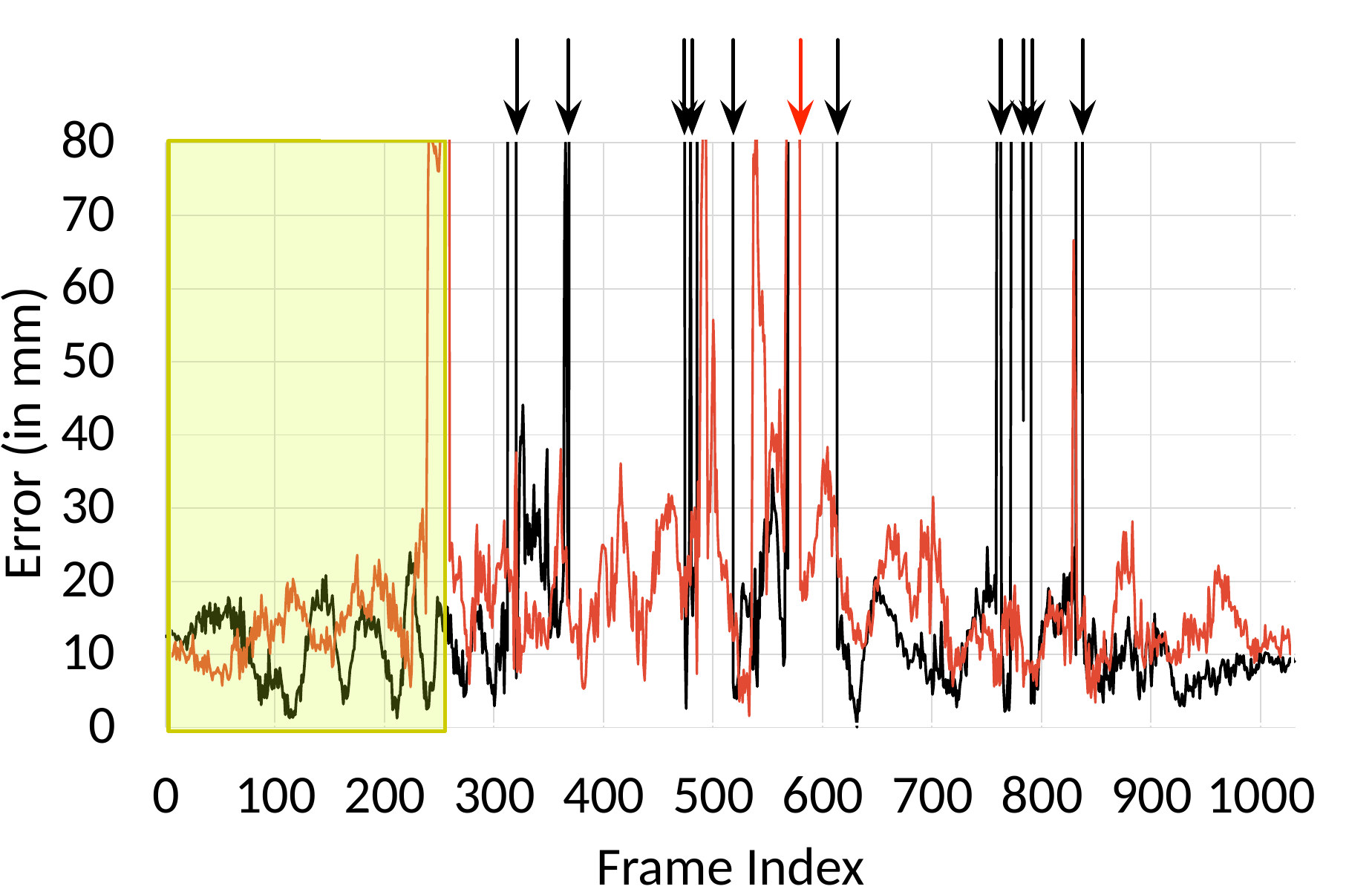} \\
(c) Casting & (d) Steamer inlay \\
\end{tabular}
\caption{Quantitative evaluation on the PROFACTOR 3D dataset and comparison with Akkaladevi et al.~\cite{akkaladevi2016tracking}. The error on all four objects from the dataset is the L2 distance between the estimated object center and the ground truth. Arrows represent frames where the trackers lose track of the objects (black for Akkaladevi et al., red for ours). The yellow overlays indicate frames that are used as training samples to fine-tune our network, where applicable. While errors are similar between the two approaches, our tracker is significantly more robust to rapid camera motions. }
\label{fig:akkaladevi-quantitative}
\end{figure}

\begin{figure*}[!t]
\centering
\footnotesize
\setlength{\tabcolsep}{1pt}
\begin{tabular}{cccccc}
%
%
%
\rotatebox{90}{\hspace{2.5em}Skull, ours} & 
\includegraphics[width=.18\linewidth]{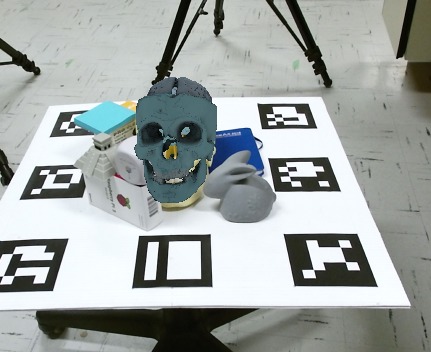} & 
\includegraphics[width=.18\linewidth]{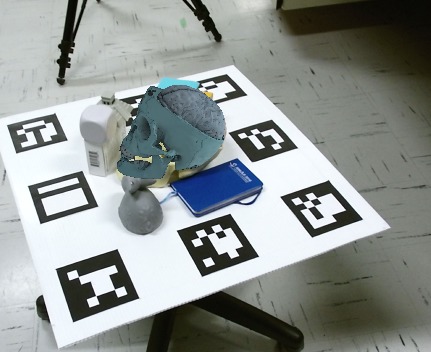} & 
\includegraphics[width=.18\linewidth]{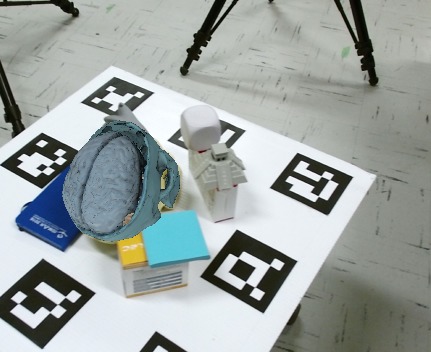} & 
\includegraphics[width=.18\linewidth]{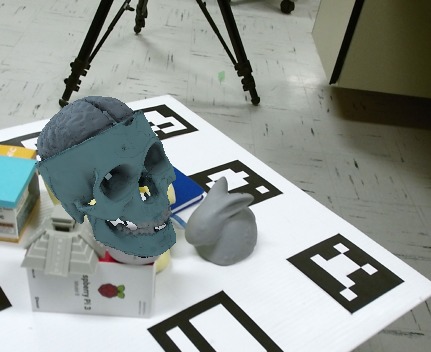} & 
\includegraphics[width=.18\linewidth]{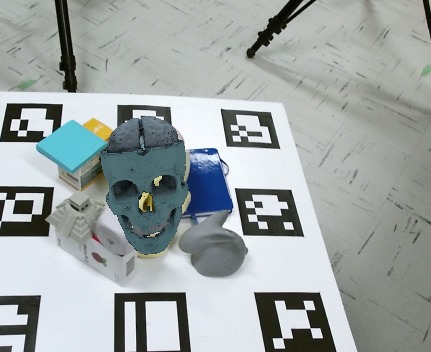} \\
\rotatebox{90}{\hspace{3em}Skull, \cite{tan-iccv-15}} & 
\includegraphics[width=.18\linewidth]{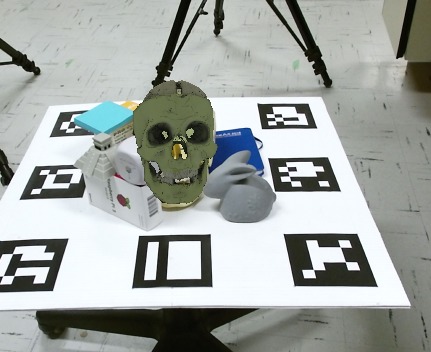} & 
\includegraphics[width=.18\linewidth]{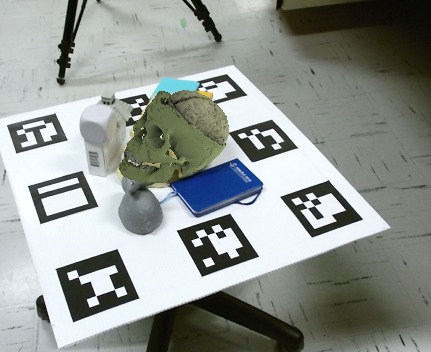} & 
\includegraphics[width=.18\linewidth]{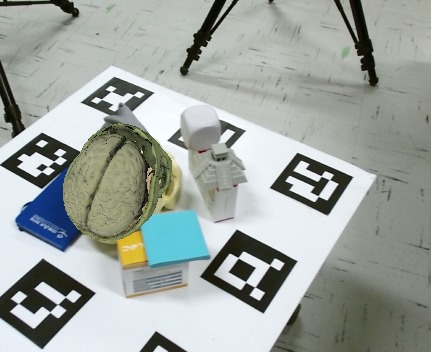} & 
\includegraphics[width=.18\linewidth]{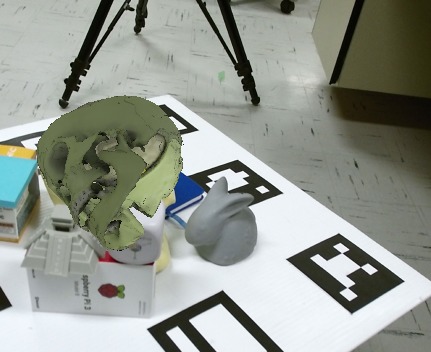} & 
\includegraphics[width=.18\linewidth]{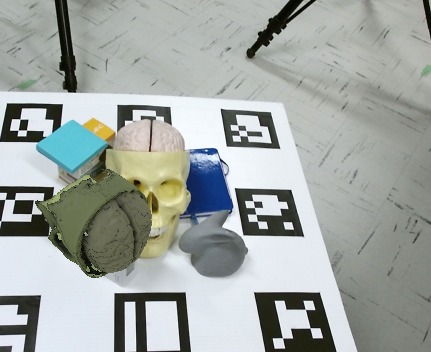} \\
\rotatebox{90}{\hspace{2.5em}Turtle, ours} & 
\includegraphics[width=.18\linewidth]{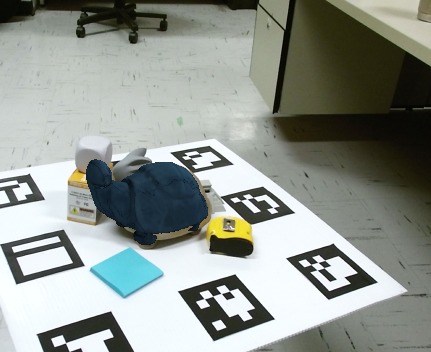} & 
\includegraphics[width=.18\linewidth]{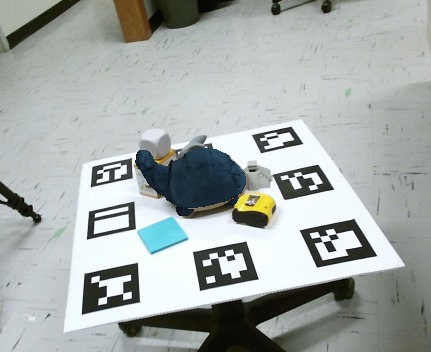} & 
\includegraphics[width=.18\linewidth]{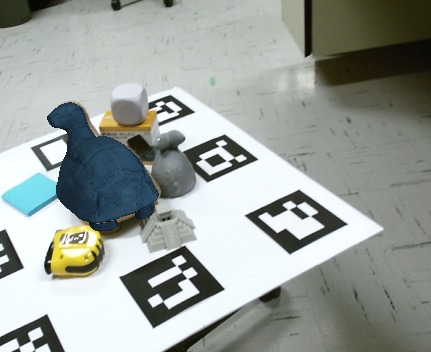} & 
\includegraphics[width=.18\linewidth]{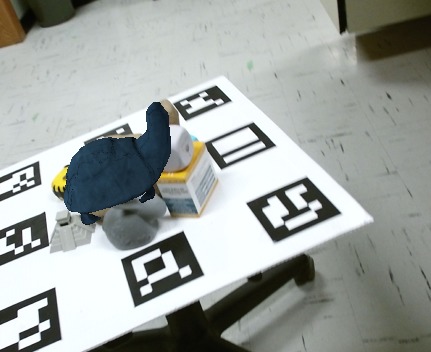} & 
\includegraphics[width=.18\linewidth]{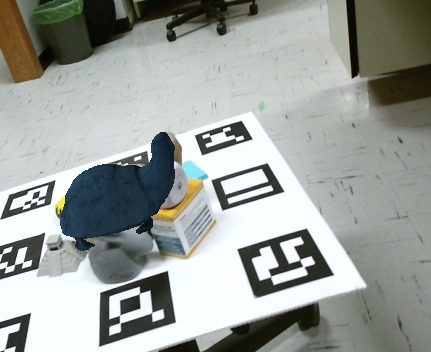} \\
\rotatebox{90}{\hspace{3em}Turtle, \cite{tan-iccv-15}} & 
\includegraphics[width=.18\linewidth]{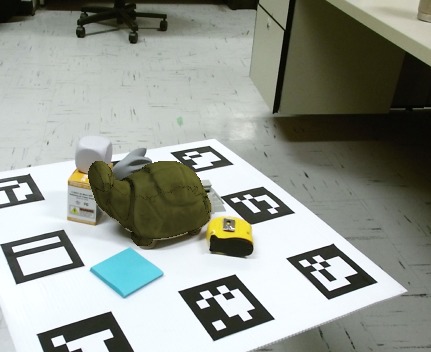} & 
\includegraphics[width=.18\linewidth]{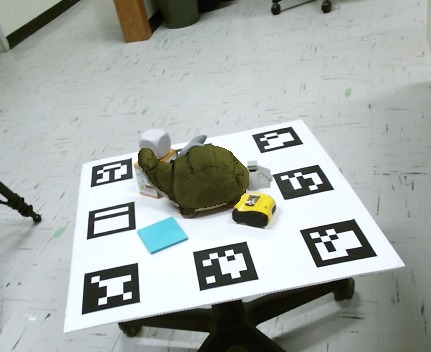} & 
\includegraphics[width=.18\linewidth]{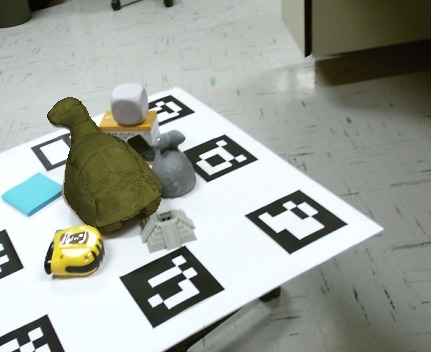} & 
\includegraphics[width=.18\linewidth]{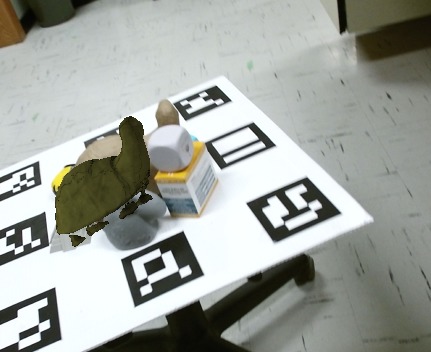} & 
\includegraphics[width=.18\linewidth]{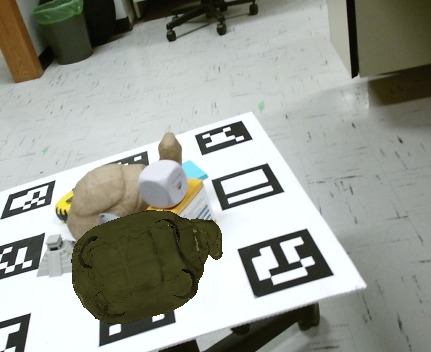} \\
%
%
%
\end{tabular}
\caption{Qualitative object tracking examples, comparing our approach with that of Tan et al.~\cite{tan-iccv-15} on the \emph{skull} and \emph{turtle} objects. Overall, our approach is more stable, more accurate, and more robust to occlusions. See supplementary material for videos of sequences from all four objects in our database: \emph{dragon}, \emph{clock}, \emph{skull} and \emph{turtle}.}
\label{fig:objects-qualitative}
\end{figure*}

\begin{table}[!t]
\centering
\begin{tabular}{lcccc}
\toprule
Object (DOF) & Tan et al.~\cite{tan-iccv-15} & Our method \\
\midrule
Dragon (translation) & 6.3 $\pm$ 3.3 mm  & \textbf{4.2} $\pm$ \textbf{2.5} mm \\
Dragon (rotation) & 1.2 $\pm$ 1.6 deg & \textbf{1.0} $\pm$ \textbf{1.0} deg \\*[.5em]
Skull (translation) & 13.3 $\pm$ 21.0 mm  & \textbf{4.6} $\pm$ \textbf{6.9} mm \\
Skull (rotation) & 17.3 $\pm$ 32.6 deg & \textbf{3.2} $\pm$ \textbf{11.1} deg \\*[.5em]
Turtle (translation) & 25.6 $\pm$ 19.2 mm  & \textbf{3.8} $\pm$ \textbf{5.9} mm \\
Turtle (rotation) & 41.4 $\pm$ 30.8 deg & \textbf{2.3} $\pm$ \textbf{3.4} deg \\*[.5em]
Clock (translation) & 26.2 $\pm$ 20.6 mm  & \textbf{3.8} $\pm$ \textbf{4.3} mm \\
Clock (rotation) & 27.9 $\pm$ 24.1 deg & \textbf{1.9} $\pm$ \textbf{3.5} deg \\*[.5em]
\bottomrule
\end{tabular}
\caption{Quantitative comparison to the approach of Tan et al.~\cite{tan-iccv-15} on four different objects.}
\label{tab:objects-quantitative}
\end{table}

\begin{figure*}[!ht]
\centering
\footnotesize
\setlength{\tabcolsep}{1pt}
\begin{tabular}{cccc}
\raisebox{.5\height}{\rotatebox{90}{\hspace{2.7em}Dragon}} & 
\raisebox{.5\height}{\includegraphics[height=2cm]{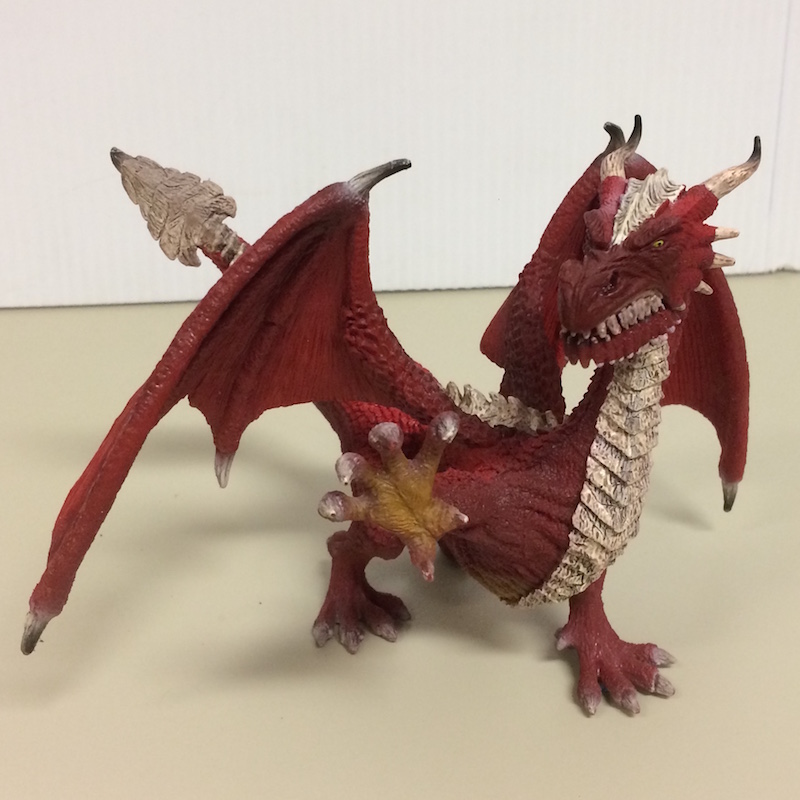}} & 
\includegraphics[width=.42\linewidth]{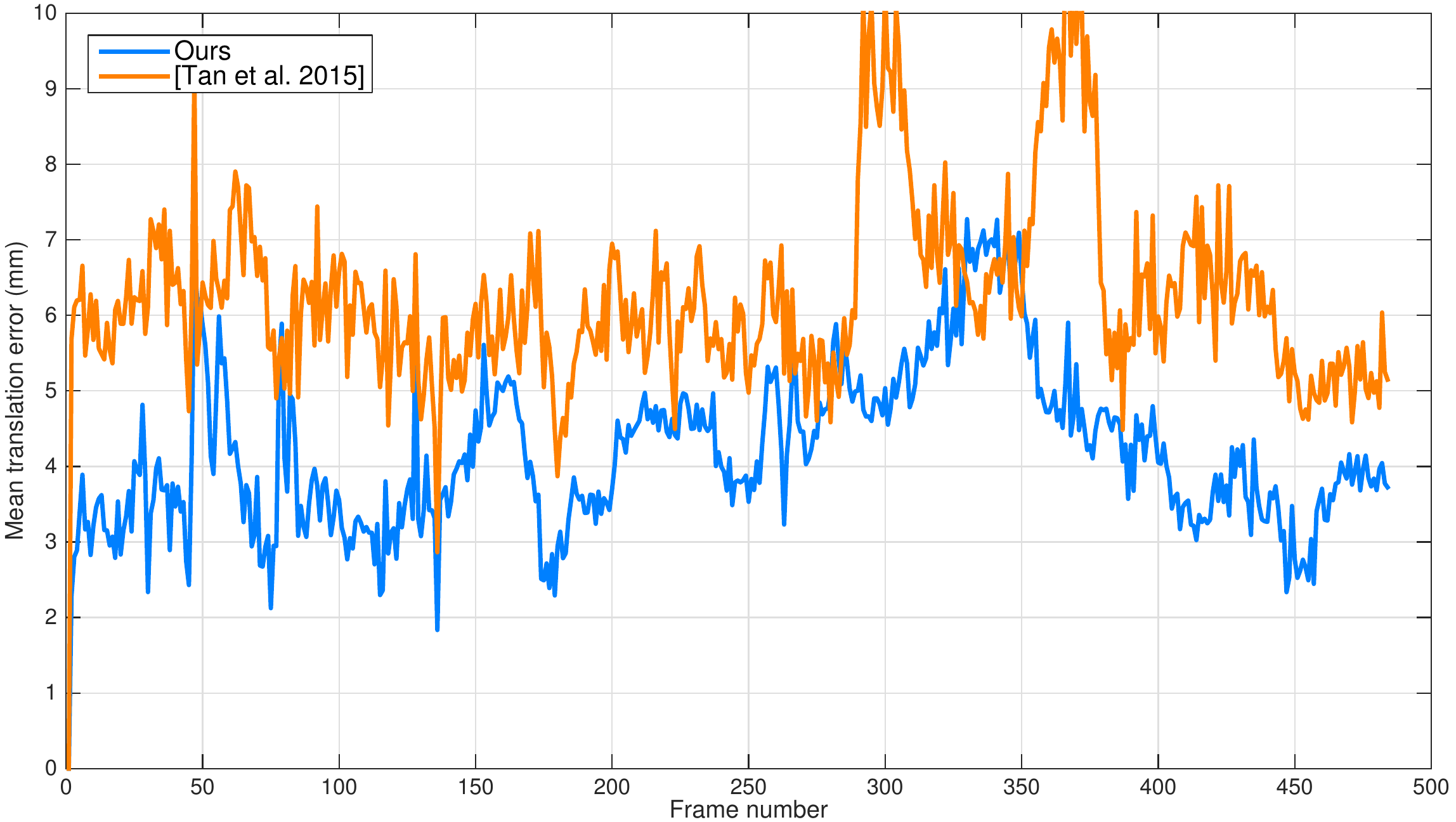} & 
\includegraphics[width=.42\linewidth]{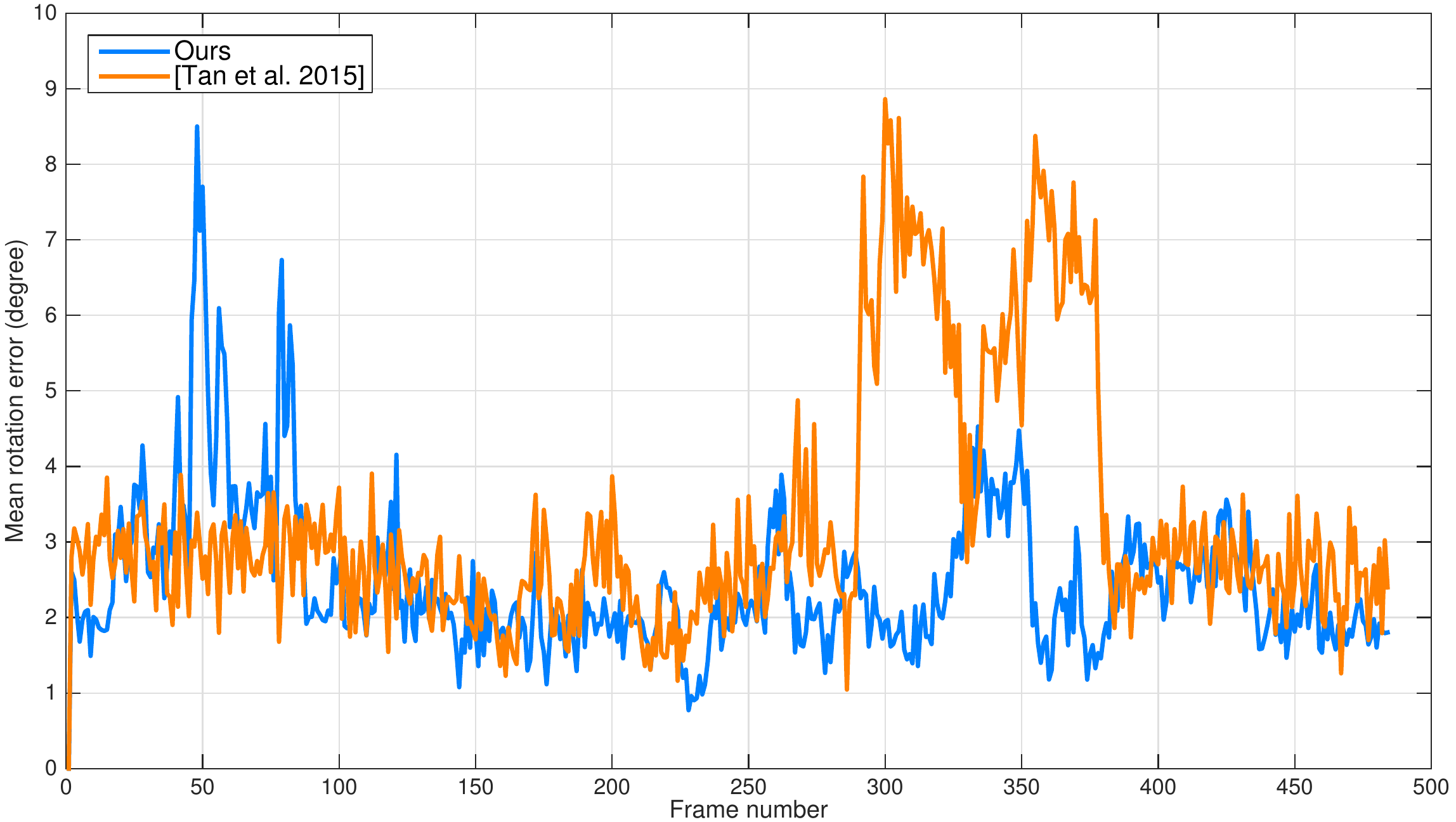} \\
\raisebox{.5\height}{\rotatebox{90}{\hspace{3.5em}Skull}} & 
\raisebox{.5\height}{\includegraphics[height=2cm]{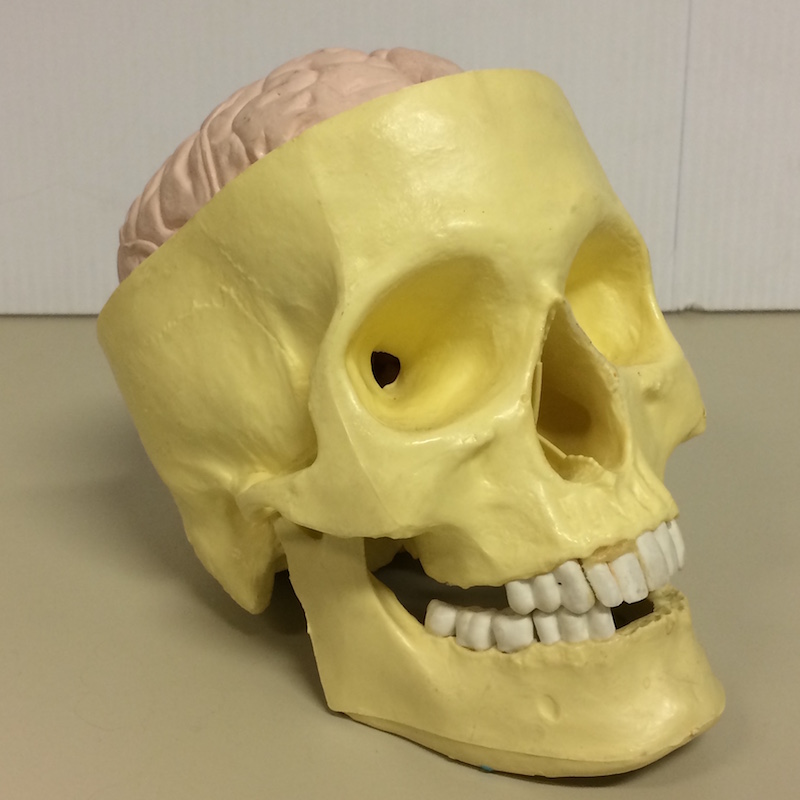}} & 
\includegraphics[width=.42\linewidth]{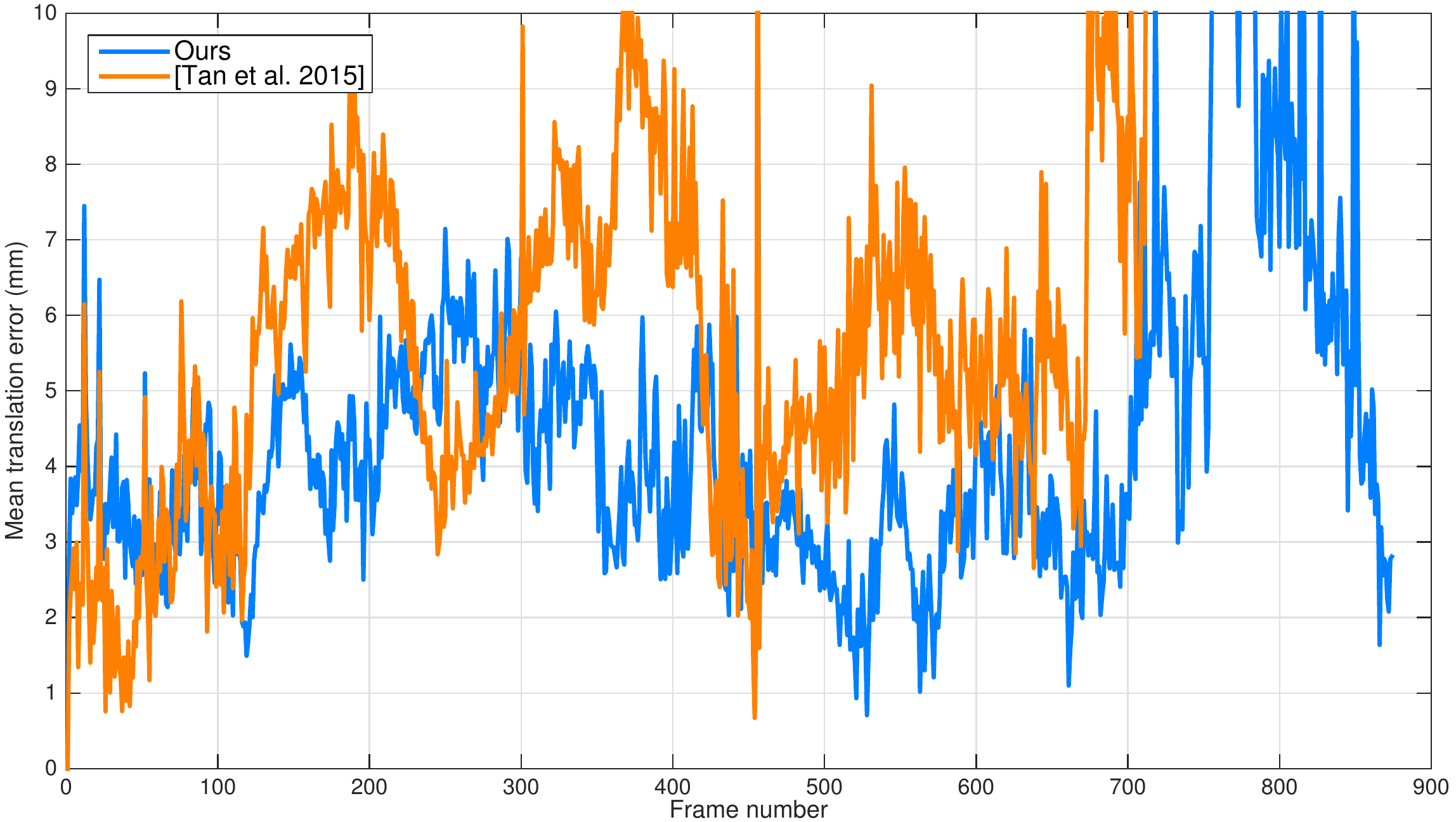} & 
\includegraphics[width=.42\linewidth]{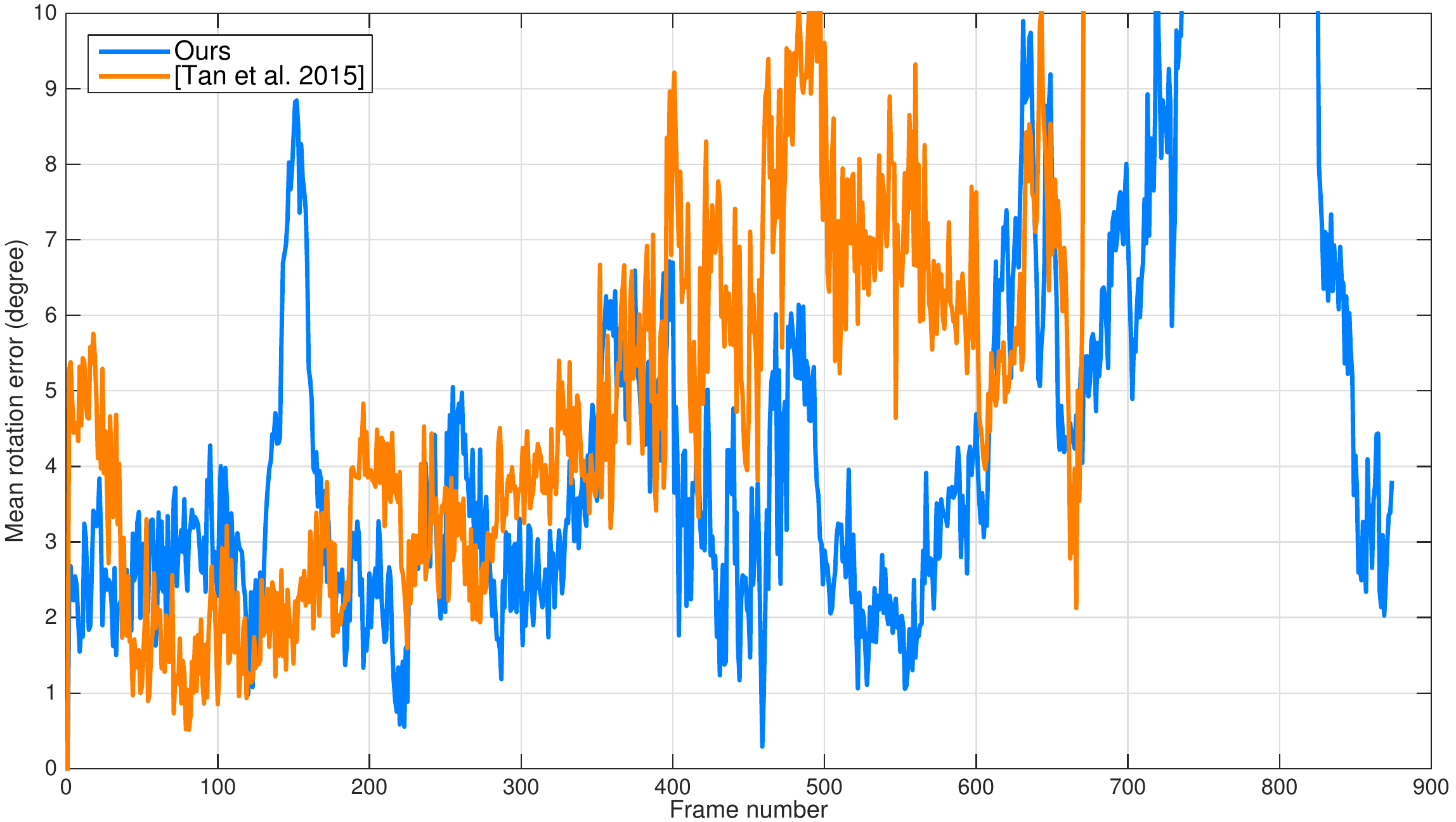} \\
\raisebox{.5\height}{\rotatebox{90}{\hspace{3em}Turtle}} & 
\raisebox{.5\height}{\includegraphics[height=2cm]{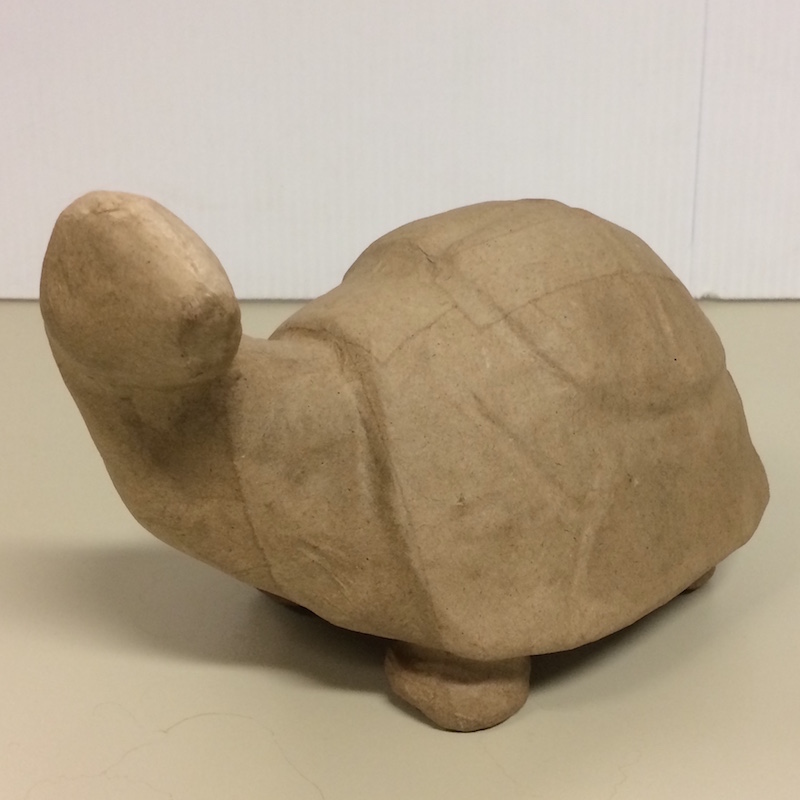}} & 
\includegraphics[width=.42\linewidth]{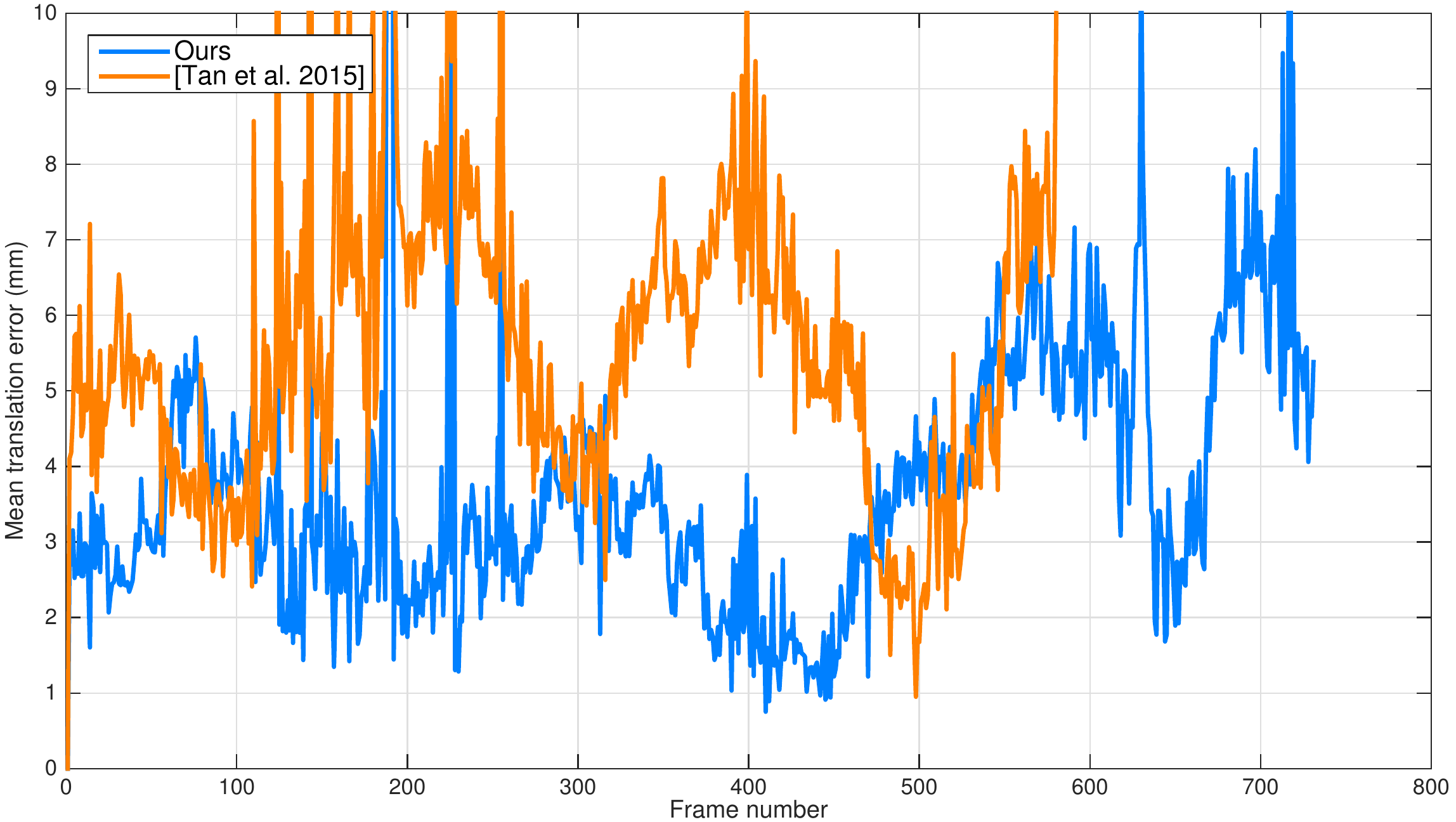} & 
\includegraphics[width=.42\linewidth]{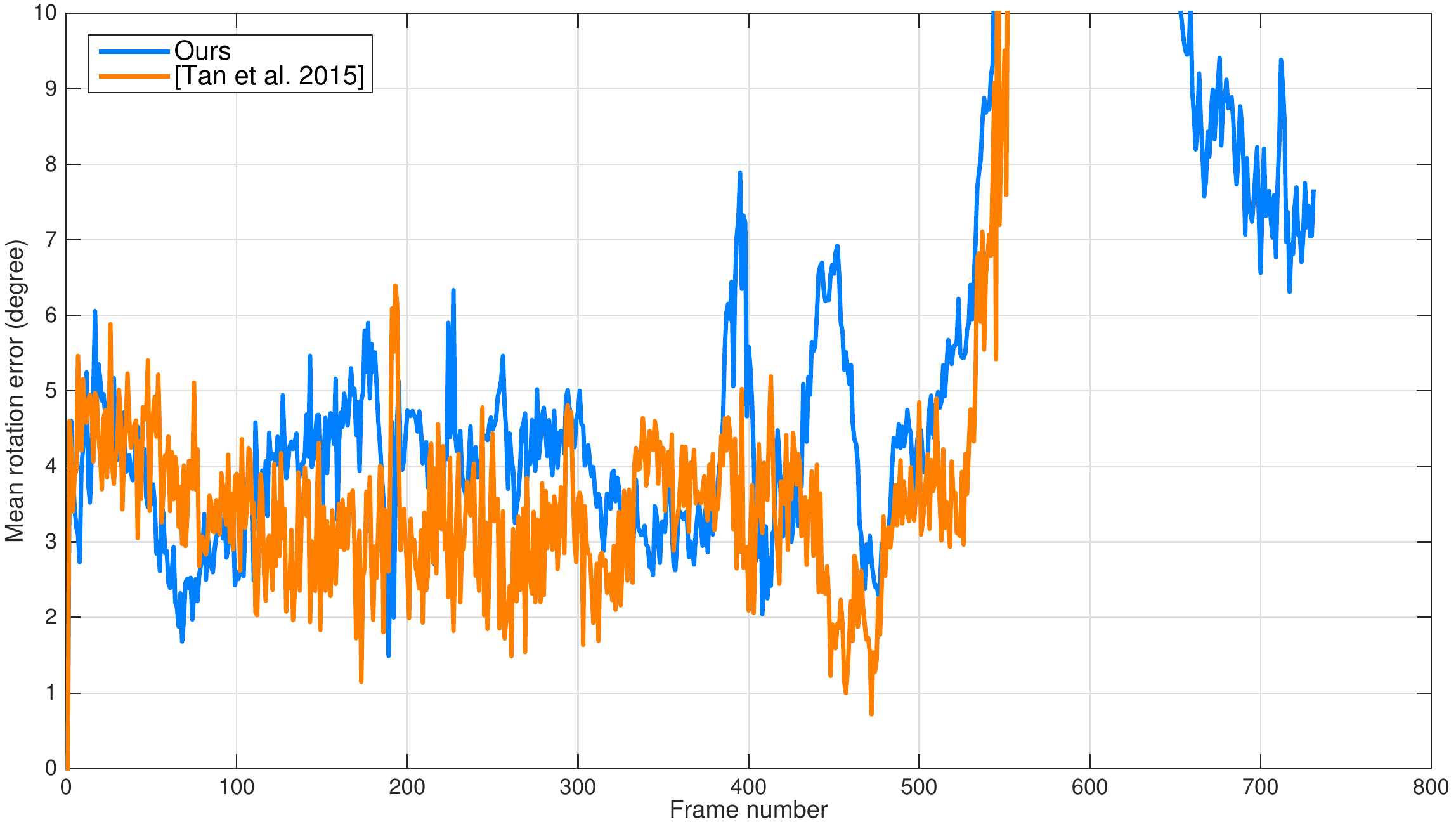} \\
\raisebox{.5\height}{\rotatebox{90}{\hspace{3em}Clock}} & 
\raisebox{.5\height}{\includegraphics[height=2cm]{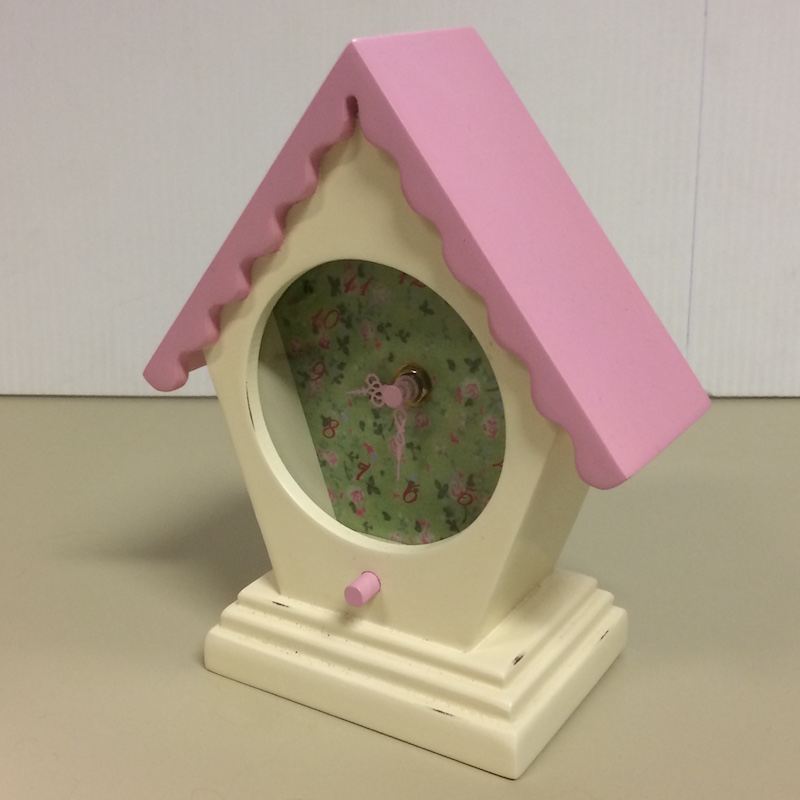}} & 
\includegraphics[width=.42\linewidth]{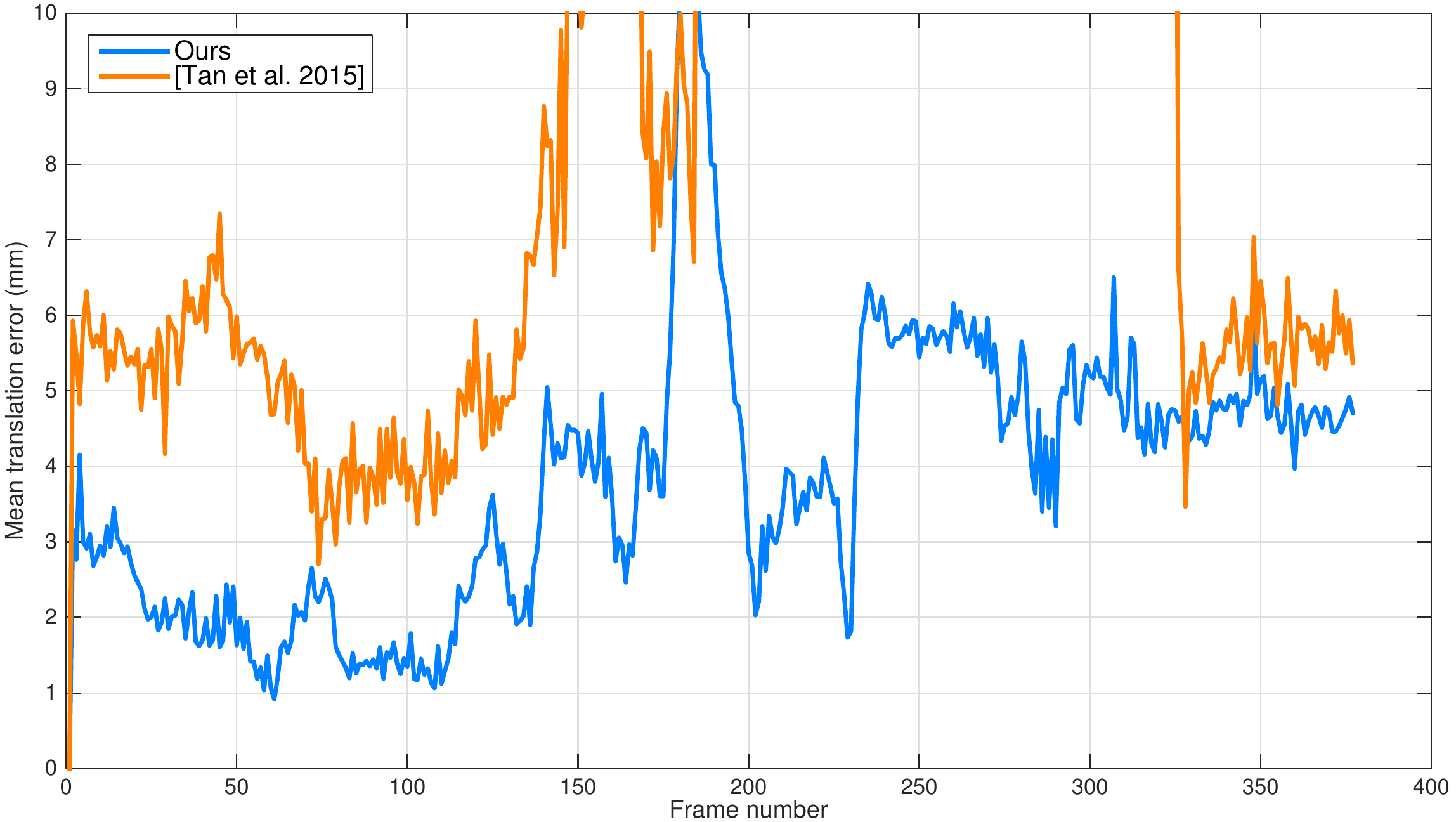} & 
\includegraphics[width=.42\linewidth]{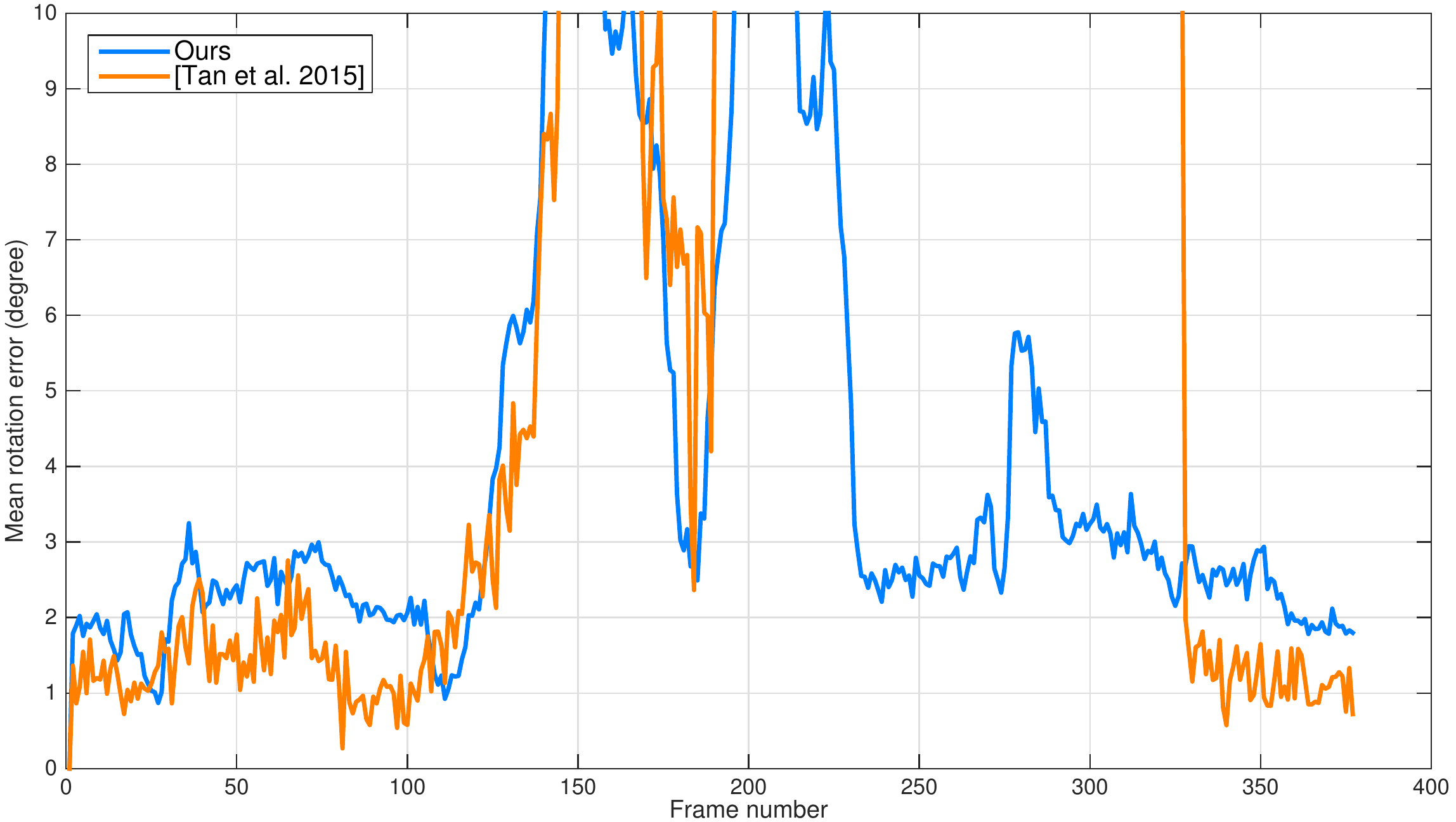} \\
& (a) Object model & (b) Translation error (in mm) & (c) Rotation error (in degrees) \\
\end{tabular}
\caption{Quantitative comparison to the approach of Tan et al.~\cite{tan-iccv-15} on 4 different objects. In each case, we plot the translation and rotation error over time for both approaches, computed against ground truth. The sequences include rotations, translations, as well as occlusions. Example images for each sequence are shown in~\autoref{fig:objects-qualitative}, and errors are summarized in~\autoref{tab:objects-quantitative}. See supplementary material for videos of these sequences. }
\label{fig:objects-quantitative}
\end{figure*}

\subsection{PROFACTOR 3D dataset}
\label{sec:dataset-of-akkaladevi}
We first compare our method on the challenging PROFACTOR 3D dataset~\cite{akkaladevi2016tracking}. This dataset provides a single sequence of a scene recorded by a fast moving Primesense camera, with four different objects placed among other clutter. The dataset benchmarks robustness to motion blur and rapid camera displacement, and contains limited occlusions overall. It also includes 3D models of the objects. The ``garden gnome'', ``casting'', and ``steamer inlay'' are reconstructed from Primesense data. The last object, ``bin'', is represented by a handmade CAD model. Note that some models have a noisy reconstruction and are missing parts and textures. We train our network using these 3D models, and, for the ``casting'' and ``steamer inlay'', we also fine-tune on the first 250 frames of the sequence since additional real data is not available. No fine-tuning was performed for the ``garden gnome'' and ``bin'' models.

We report the distance of the prediction's center with the ground truth in~\autoref{fig:akkaladevi-quantitative} for each object, and compare the performance with that of the method of Akkaladevi et al.~\cite{akkaladevi2016tracking}. In addition to pose estimation errors indicated by the curves (black for Akkaladevi et al.~\cite{akkaladevi2016tracking}, red for ours), we also report points at which the trackers lose track of the objects and must be reset. These frames are indicated by arrows. Note that when our tracker completely loses track of the object, it is reinitialized with the ground truth pose. Overall accuracy is similar for both methods on all four objects. However, we note that our method is more robust to rapid camera motion (and resulting motion blur) since our tracker can recover from large errors without having to be reinitialized. For example, the ``garden gnome'' is never lost and the ``bin'' is lost only once, compared to 1 and 5 times for Akkaladevi et al., respectively (top row of~\autoref{fig:akkaladevi-quantitative}). The more challenging objects, ``casting'' and ``steamer inlay'', are lost 2 and 1 times by our method compared to 6 and 10 times respectively by Akkaladevi et al. (bottom row of~\autoref{fig:akkaladevi-quantitative}). This experiment shows that our method obtains competitive overall accuracy while being more robust to fast camera motion, even in the absence of quality data for training.

\subsection{Dataset Acquisition}
\label{sec:real-data-acquisition}

Many of the previous studies on RGBD temporal tracking compare their approaches on a dataset of synthetic images~\cite{Choi2013}, which consists in 4 objects in a virtual environment without camera noise or any type of imperfections. We strongly believe this dataset to be insufficient to encourage progress in this area, as the resulting images are completely unrealistic. Depth cameras are known for being noisy, and real environments have a richness that is hard to emulate with virtual models. The most popular dataset for evaluating 6-DOF object \emph{detection} is the Linemod dataset~\cite{hinterstoisser2012model}. However, it does not contain sequences of continuous object motions; rather it is a series of images of an object with known, but seemingly unrelated, poses. The CoRBS dataset~\cite{wasenmuller2016corbs} is also available, but focuses on evaluating SLAM algorithms. As such, it contains large static objects which do not move independently from the camera.

\paragraph{Setup} To benchmark our approach, we acquire a novel dataset of sequences of continuous object and camera motions, where the ground truth object pose with respect to the camera is known. Fiducial markers~\cite{Aruco2014} are used to obtain a reference frame. With this reference, multiple frames are fused together and ICP is used to match the reconstructed object with its 3D model. We manually finetune the object reference coordinate system from the ICP match to obtain the transform between the fiducial marker and the object coordinate system. The data is captured with a Microsoft Kinect 2.0. Example frames obtained with this approach are shown in~\autoref{fig:objects-qualitative} and~\autoref{fig:occlusions-qualitative}.

\paragraph{Dataset description} For every object, a handheld sequence with various clutter and occluding objects is provided to benchmark general accuracy and robustness. We also capture a series of video sequences where the object is placed on a turntable, and a static occluder is placed between the camera and the object. Videos are captured with the occluder placed on the side (vertical occlusion) and on the top (horizontal occlusion) of the object. One video is captured for 10\%, 20\%, 30\%, and 40\% occlusion, in both the vertical and horizontal case. Our dataset is publicly available\footnote{See {\scriptsize \url{http://www.jflalonde.ca/projects/deepTracking}}.} to encourage further research in this area.

\paragraph{Training Data} This setup also enables us to capture data with ground truth labels used to fine-tune the model (see~\autoref{sec:training-details}). The ground truth label defines the object mask for all frames and can be used to segment the background and replace it with a more realistic one~\cite{xiao2013sun3d}. Note that the fiducial markers are never seen in the training dataset. To augment the dataset, we randomly rotate the object around the camera roll axis. We captured 180 viewpoints for training and 60 viewpoints as validation on the top hemisphere only. Naturally, these fine-tuning frames are used exclusively for training: testing is performed on an altogether separate set of video sequences.

\subsection{Quantitative comparison using real objects}

We use our real dataset to compare against Tan et al.~\cite{tan-iccv-15} in a quantitative way by reporting the mean error in translation and rotation. Since no public code is available, we implemented their algorithm following the paper directions. To ensure fairness, our implementation was validated on the synthetic dataset of Choi et al.~\cite{Choi2013} and resulted in performances similar to those reported in the original paper. Note that we do not include a comparison with ICP as Tan et al. already demonstrated that their method outperforms ICP in terms of accuracy, robustness and speed.

We report results in three different ways: by showing representative frames in~\autoref{fig:objects-qualitative}, by plotting the translation and rotation error over time for each sequence in~\autoref{fig:objects-quantitative}, and by reporting the mean error in~\autoref{tab:objects-quantitative}. The ``dragon'' sequence is relatively easy because it does not contain very severe occlusions. Both methods report low errors of $6.26$mm and $4.19$mm in translation, and $1.24^\circ$ and $1.06^\circ$ in rotation for Tan et al. and ours, respectively. While ours reports lower numbers, both yield qualitatively very similar results (see supplementary video). 

More important differences arise when occlusion is more severe. In the three other videos (``skull'', ``turtle'', and ``clock''), both methods fare well in the early frames, but react differently when occlusion occurs. Typically, Tan et al. suffer from irrecoverable failure, and tracking is lost (although sometimes it catches back on, as in the ``clock'' sequence). In contrast, our method is able to maintain relatively low error and continues to track the object.

\subsection{Robustness to occlusion} 
\label{sec:robustness-occlusion}

We systematically evaluate the robustness of our approach to occlusions, and compare it to Tan et al.~\cite{tan-iccv-15} on our dataset of controlled occlusion cases. To ensure that each tracker is not artificially penalized for missing a single frame (and thus reporting very high error for this single mistake), each tracker is initialized to the ground truth pose every 15 frames. 

~\autoref{fig:occlusions-quantitative} reports the errors on translation and rotation for all occlusion scenarios. Both methods perform similarly when occlusion is 10\%. In the 20\% case, differences begin to arise: the translation error of Tan et al. begins to increase, most notably in the horizontal occluder scenario. Differences continue to increase with occlusion percentage: when it reaches 30\%, the approach of Tan et al.~\cite{tan-iccv-15} often loses track of the object almost immediately, while ours maintains a reasonable estimate. These results are corroborated by example frames from selected sequences, shown in~\autoref{fig:occlusions-qualitative}. The horizontal occluder case appears to be more difficult for both methods: this is probably due to the fact that the dragon head and wings represent discriminative features that are helpful for tracking. See the supplementary material for all the videos. 

\begin{figure}[!t]
\centering
\footnotesize
\setlength{\tabcolsep}{1pt}
\begin{tabular}{cc}
\includegraphics[width=.49\linewidth]{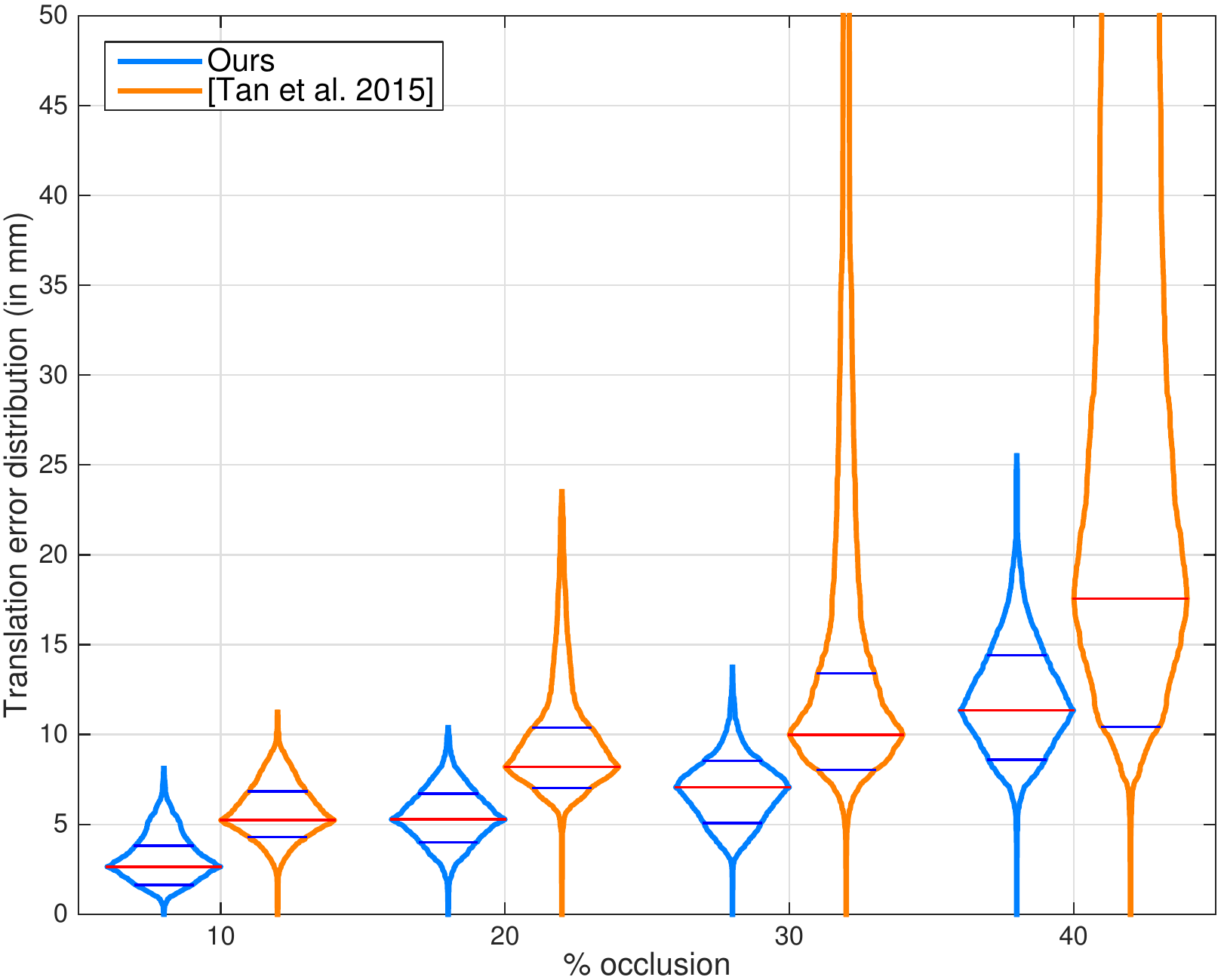} &
\includegraphics[width=.49\linewidth]{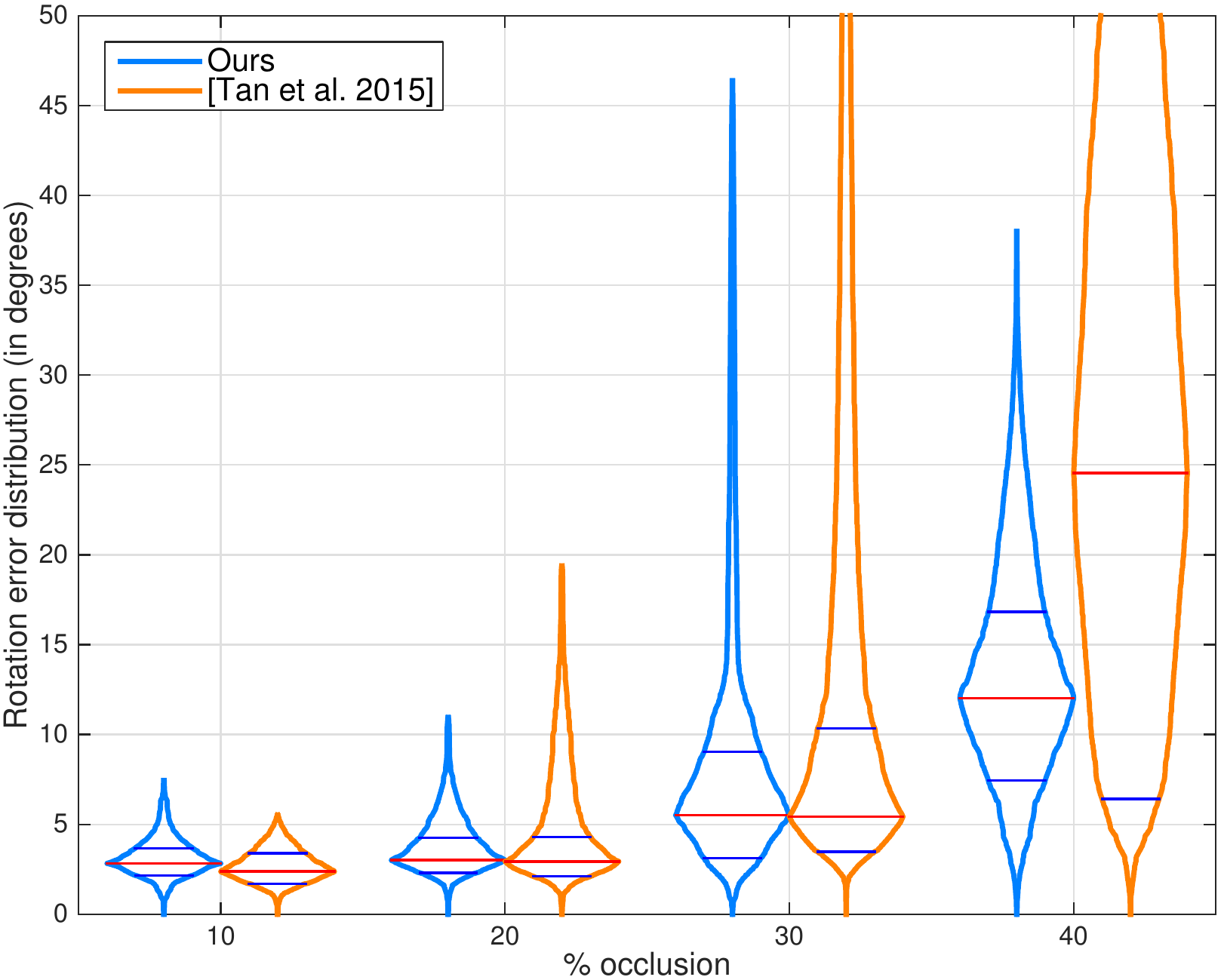} \\
\multicolumn{2}{c}{(a) Vertical occluder} \\*[1em]
\includegraphics[width=.49\linewidth]{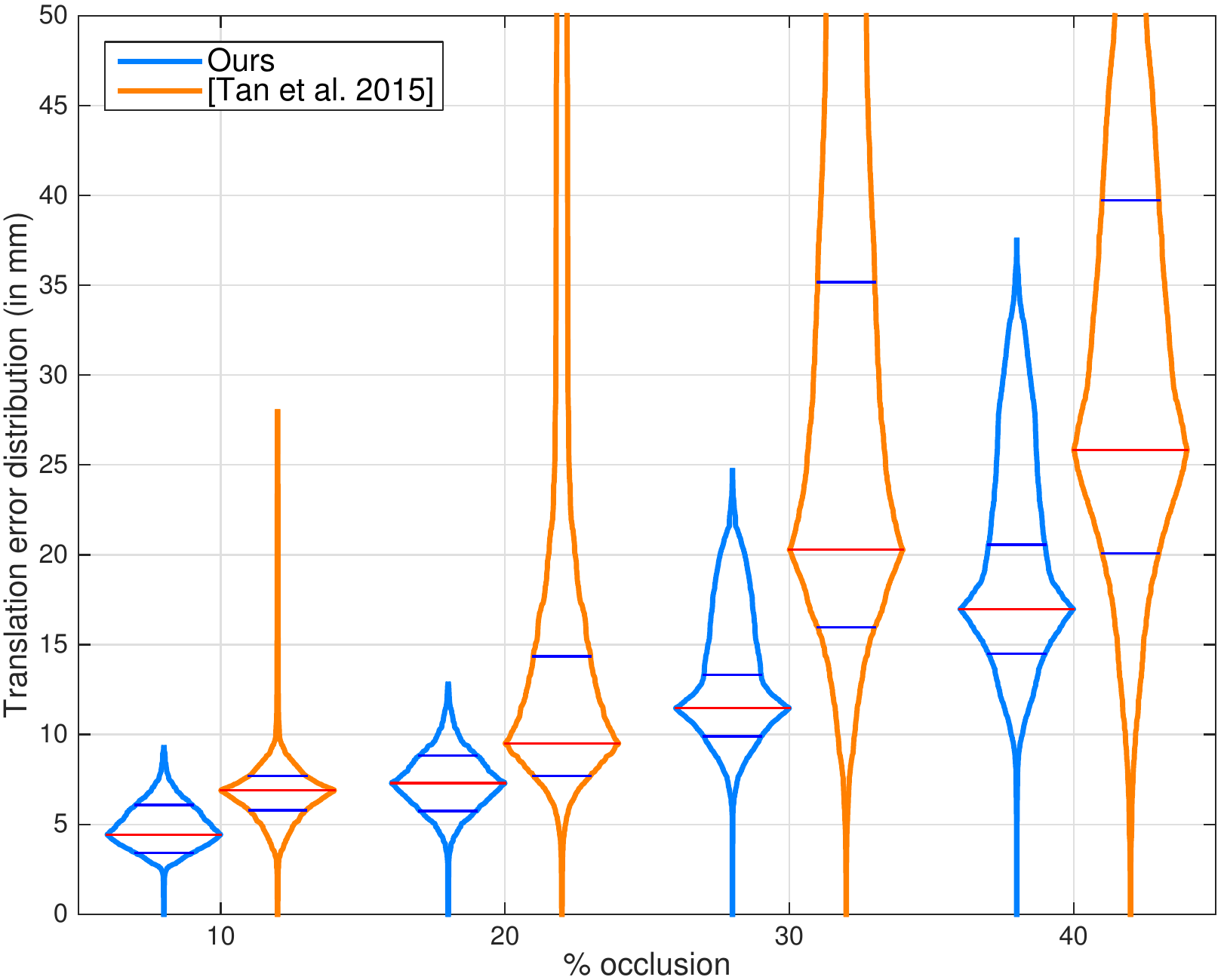} &
\includegraphics[width=.49\linewidth]{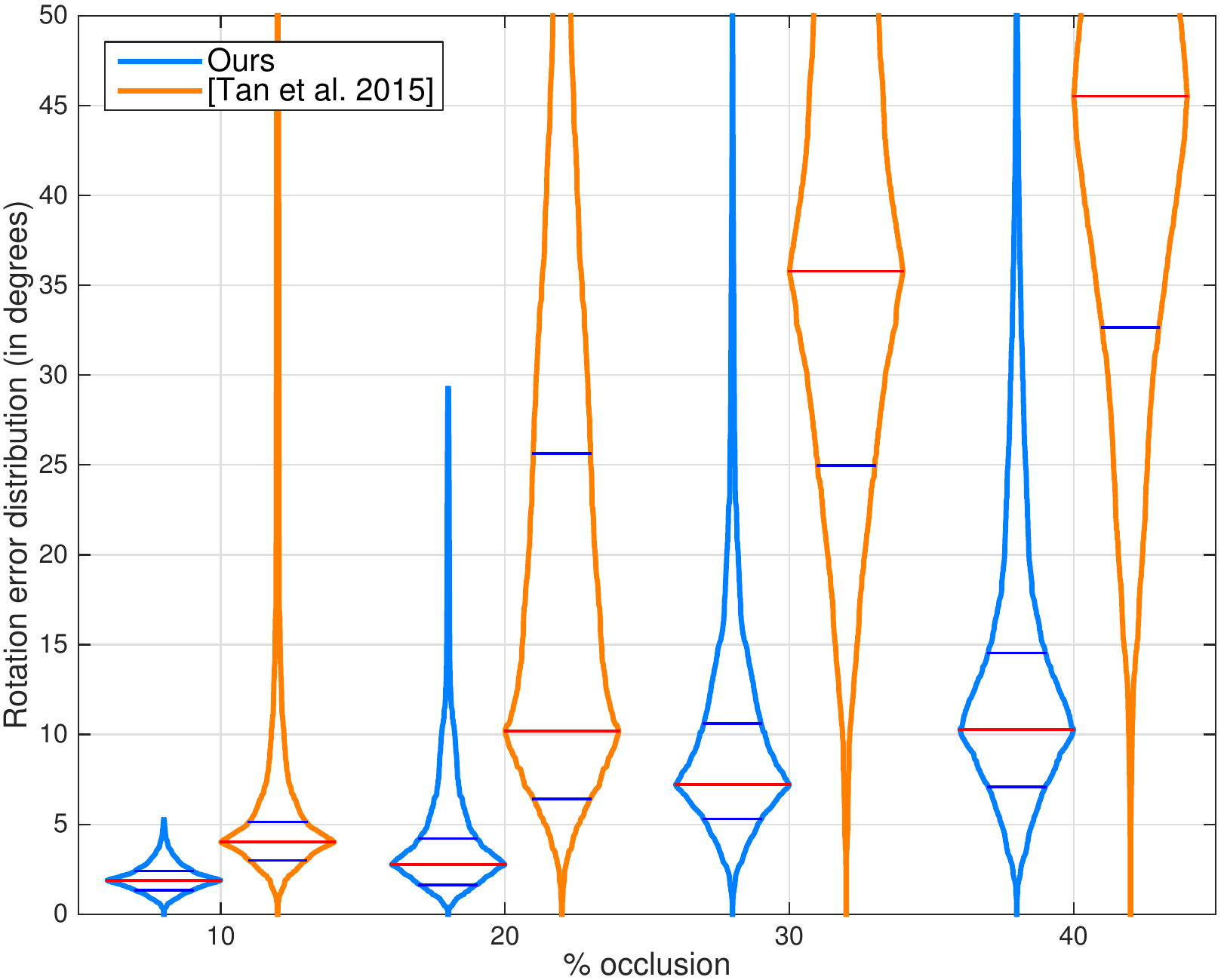} \\
\multicolumn{2}{c}{(b) Horizontal occluder} \\
\end{tabular}
\caption{Quantitative evaluation of robustness to occlusions, comparing our approach to that of Tan et al.~\cite{tan-iccv-15}. The object is placed on a turntable and occluded by a static (a) vertical and (b) horizontal occluder in increasing amounts, ranging from 10\% to 40\%. To display the errors, we use ``box-percentile plots''~\cite{esty2003box}, which illustrate the distribution of errors vertically. The red horizontal bars indicate the median, while the bottom (top) blue bars are the 25th (75th) percentiles. See supplementary material for videos of these sequences. }
\label{fig:occlusions-quantitative}
\end{figure}

\subsection{Robustness to initialization}
\label{sec:initialization}

In the experiments presented above, the tracker is always initialized to the ground truth object position. Since that position may not be known in a real-life scenario (where the tracker would, for example, be initialized by another algorithm~\cite{akkaladevi2016tracking}), we also provide an analysis of the sensitivity of the tracker to perturbations in initialization and show results in~\autoref{fig:initialization-quantitative}. To perform this analysis, 10 frames are randomly selected from the ``dragon'' sequence, and random perturbations of increasing magnitude are applied to the initial pose given to the tracker. The tracker is then iterated 15 times, and the L2 error between the final predicted pose and the ground truth is computed. For clarity, perturbations in rotation and translation are computed independently. The rotation perturbations range from 5 to $75^\circ$ in increments of $5^\circ$, and 10 to 130mm in increments of 10mm in translation. Random perturbations are sampled 40 times for each increment, and the mean and standard deviation of the error are reported in~\autoref{fig:initialization-quantitative}.~\autoref{fig:initialization-quantitative}-a) presents the rotation and translation error in the presence of rotation perturbations. The plot shows that after a magnitude of $35^\circ$, the rotation error increases while the translation error stays low.~\autoref{fig:initialization-quantitative}-b) shows the error with translation perturbations. In this case, both rotation and translation errors increase gracefully with a perturbation distance greater than 90mm. It is worth noting that, since the object is roughly 150mm in size, such an error in initialization results in a bounding box which overlaps the object by less than 50\%. Our method can thus handle noisy initialization and still converge very close to the ground truth position.

\subsection{Speed and memory footprint}

At test time, two main steps must be accomplished: rendering the predicted frame $\mathbf{x}_\text{pred}$, and estimating the relative pose of the object using the CNN. Rendering takes less than 1ms, and inference with the CNN takes 7ms, for a total of 8ms per frame. These timings were computed on an Nvidia GTX970M (mobile) GPU using a python implementation. Because of its simplicity, the memory footprint of the entire network is a mere 145 MB.

\section{Conclusion}

We have presented a new method for 6-DOF tracking of rigid objects leveraging deep neural networks. Using an end-to-end learning method enables the network to automatically learn features adapted to the task from data. Using learned features enables both robustness to higher amount of occlusions and also improves overall tracking accuracy. In addition, we also introduce a new temporal tracking dataset that will be available publicly. This dataset fills the gap for availability of real data tracking benchmarks and offers a novel method to quantify robustness against occlusion. We thoroughly compare our approach to the state-of-the-art by Tan et al.~\cite{tan-iccv-15} and Akkaladevi et al.~\cite{akkaladevi2016tracking} and find that, in almost all cases, our approach outperforms them in terms of accuracy and robustness to occlusions and to fast camera motion.

A current limitation of our approach is that the model learns only a single object at a time. Therefore, part of the learning process must be restarted for each object. However, multiple models can be computed and rendered in parallel on the GPU. Research has shown that neural networks can be good at generalization~\cite{wohlhart2015learning}, an interesting future research direction would be to apply these generalization techniques to the temporal tracking scenario.

Training deep neural networks is significantly more computationally intensive than the random forests used in \cite{tan-iccv-15}. Early experiments seem to indicate that generic, pre-trained models obtained on various 3D models can be used to bootstrap object-specific models. This could potentially improve training speed. Finally, while our method performs remarkably well in the presence of occlusion, it may still suffer from tracking failures. Another interesting future direction would be to produce confidence about the output and bridge the gap between detection and temporal tracking.

\begin{figure}[!th]
\centering
\footnotesize
\setlength{\tabcolsep}{1pt}
\begin{tabular}{cc}
 \includegraphics[width=.49\linewidth]{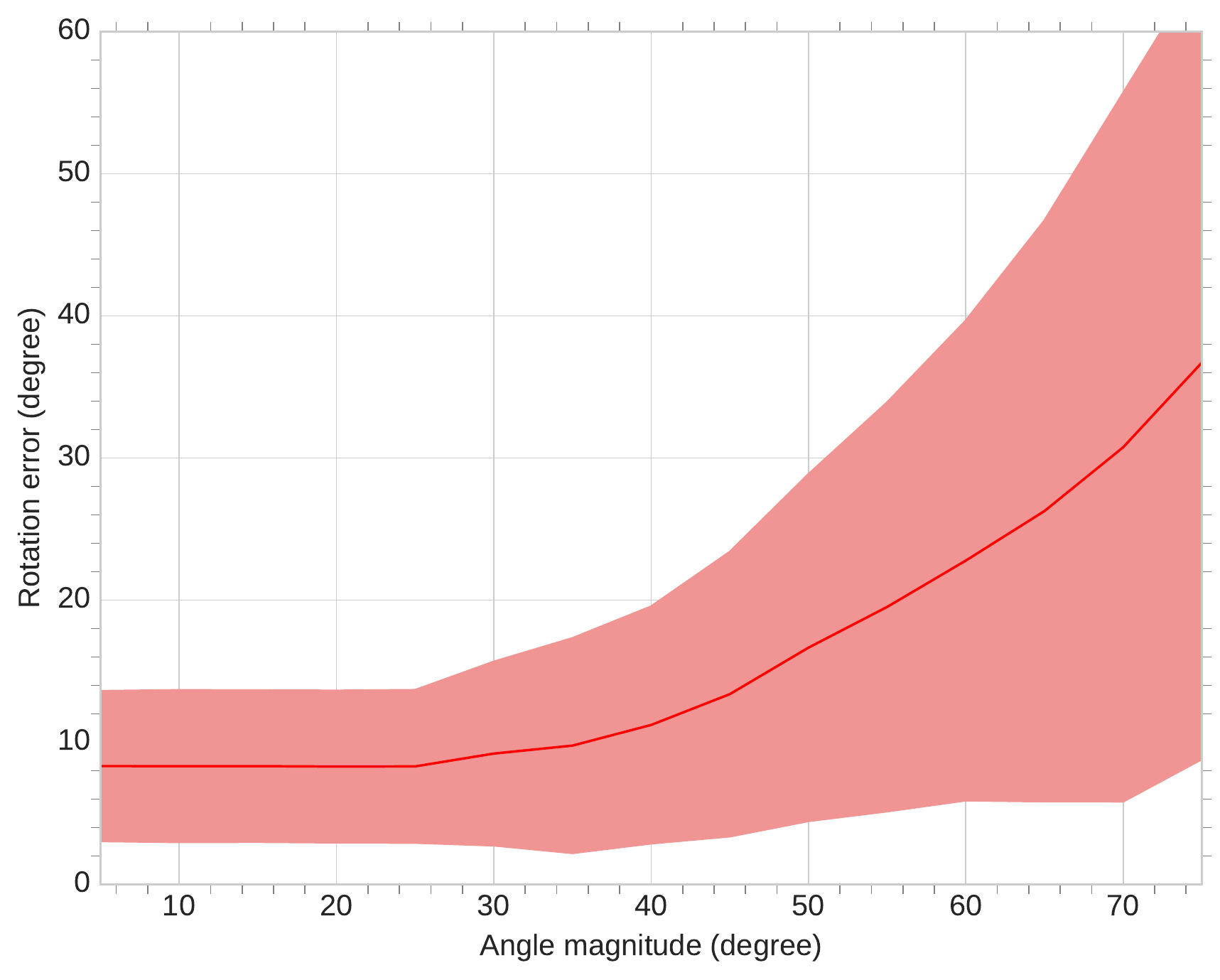} &
 \includegraphics[width=.49\linewidth]{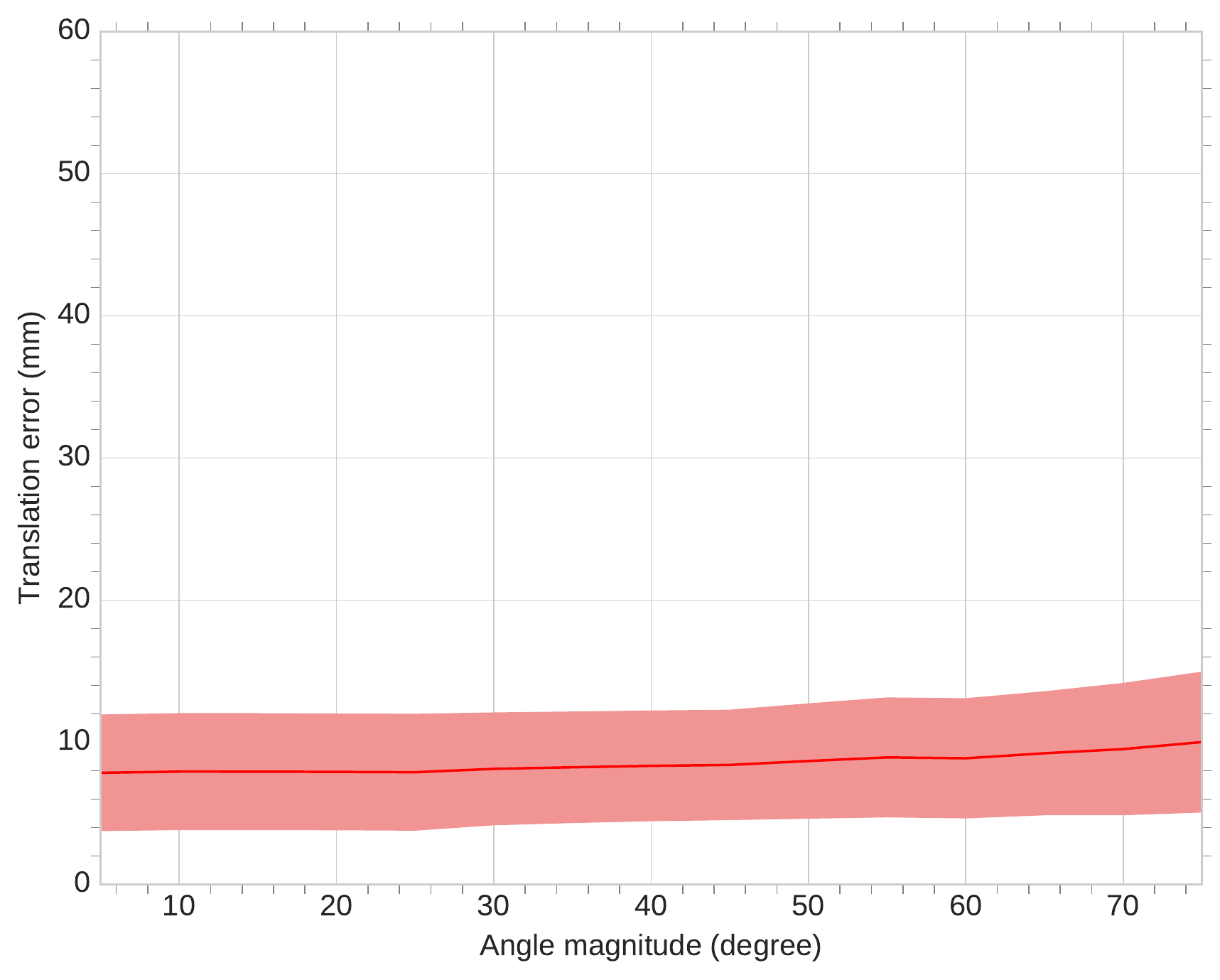} \\
 \multicolumn{2}{c}{(a) Rotation noise} \\*[1em]
 \includegraphics[width=.49\linewidth]{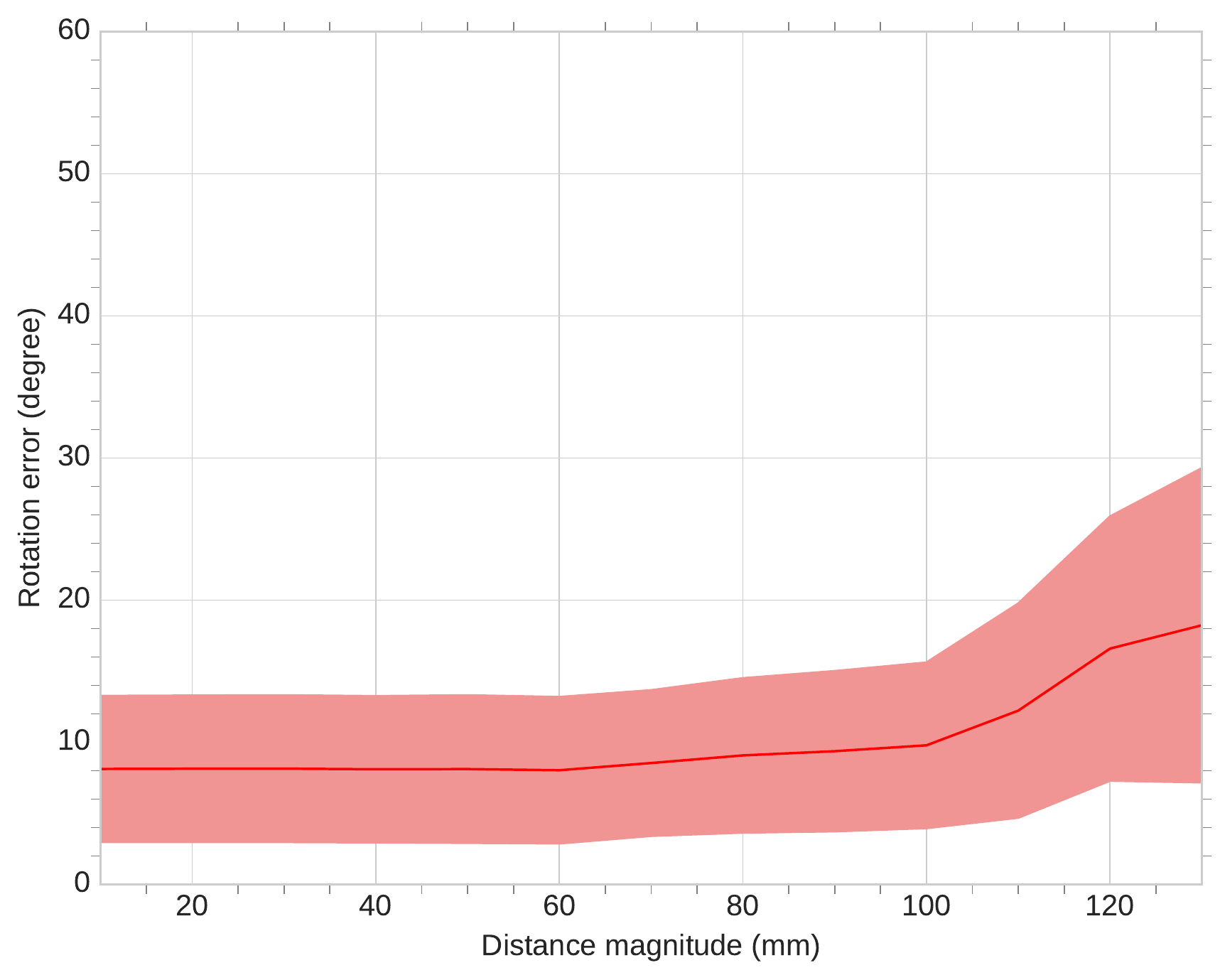} &
 \includegraphics[width=.49\linewidth]{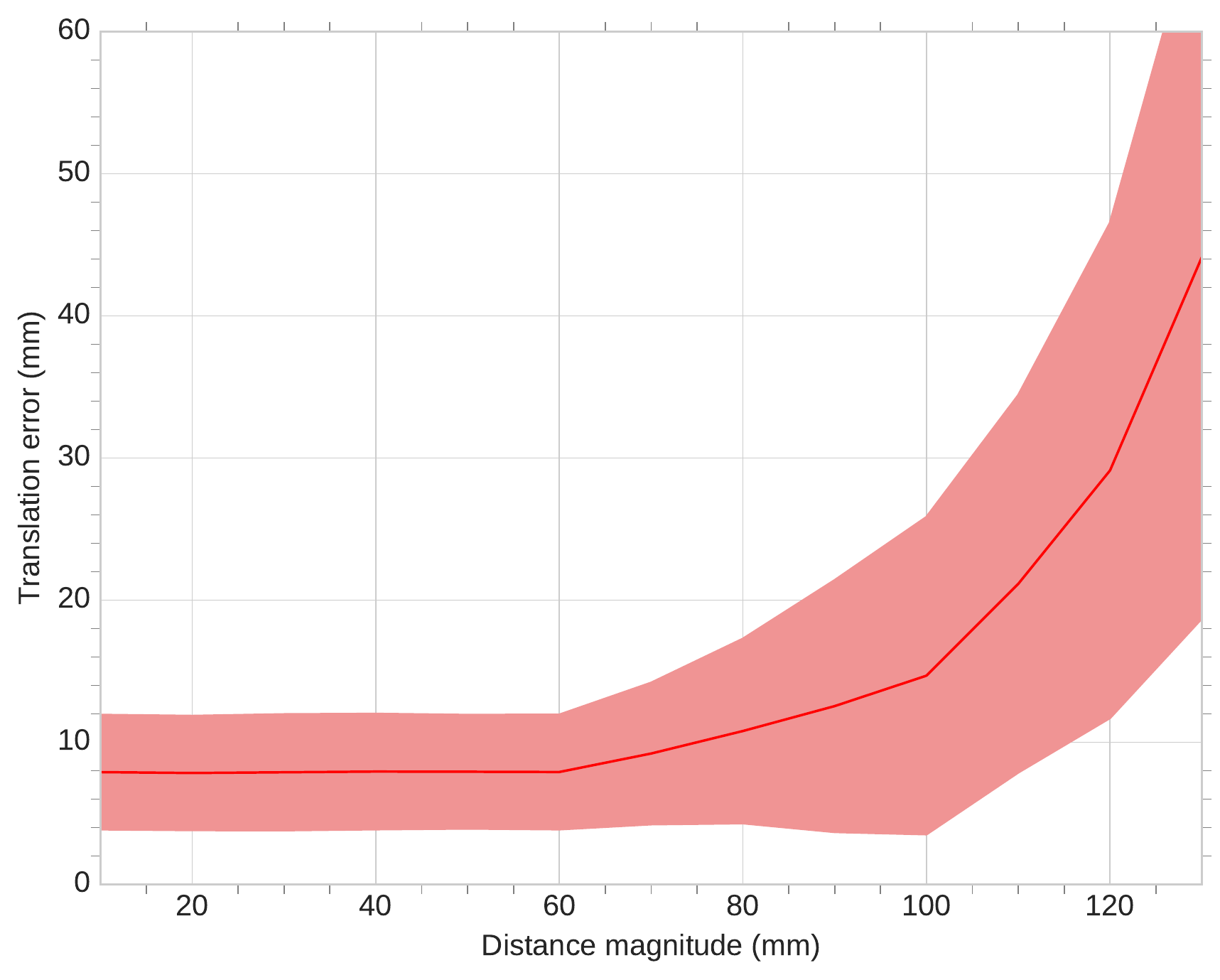} \\
 \multicolumn{2}{c}{(a) Translation noise} \\*[1em]

\end{tabular}
\caption{Quantitative evaluation of robustness to perturbations in initialization. (a) The L2 norm of the (left) rotation and (right) translation error between the network prediction and the ground truth is plotted against the magnitude of random perturbations in rotation. (b) The same errors are plotted against perturbations in translation. Note that we apply 15 iterations of the tracker before computing the errors. The red line is the mean error and the margin is the standard deviation. Overall, our tracker is robust to large errors in initial pose, and fails gracefully beyond translation errors of roughly 90mm.}
\label{fig:initialization-quantitative}
\end{figure}

\begin{figure*}[!t]
\centering
\footnotesize
\setlength{\tabcolsep}{1pt}
\begin{tabular}{cccccc}
\rotatebox{90}{\hspace{1em}20\% occlusion, ours} & 
\includegraphics[width=.19\linewidth]{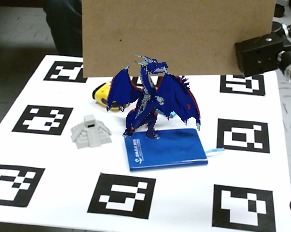} &
\includegraphics[width=.19\linewidth]{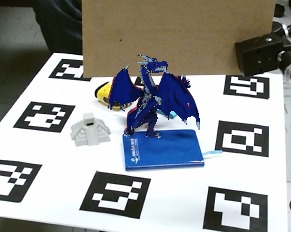} &
\includegraphics[width=.19\linewidth]{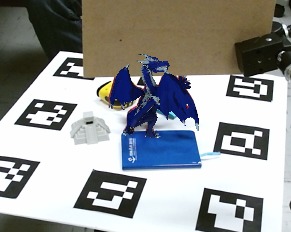} &
\includegraphics[width=.19\linewidth]{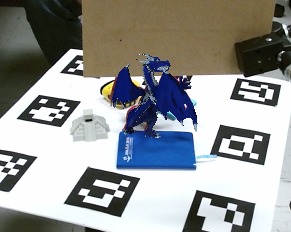} &
\includegraphics[width=.19\linewidth]{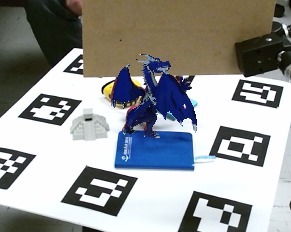} \\
\rotatebox{90}{\hspace{1em}20\% occlusion, \cite{tan-iccv-15}} & 
\includegraphics[width=.19\linewidth]{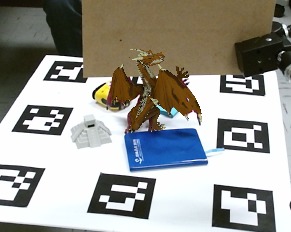} &
\includegraphics[width=.19\linewidth]{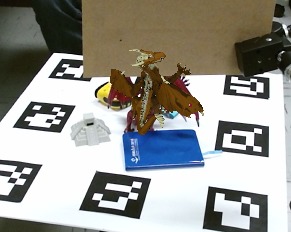} &
\includegraphics[width=.19\linewidth]{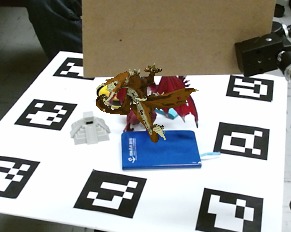} &
\includegraphics[width=.19\linewidth]{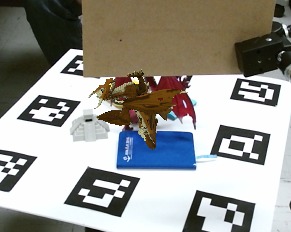} &
\includegraphics[width=.19\linewidth]{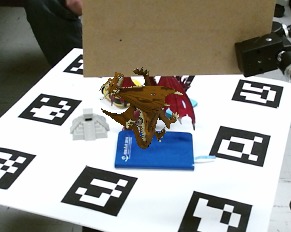} \\*[1em]
\rotatebox{90}{\hspace{1em}40\% occlusion, ours} & 
\includegraphics[width=.19\linewidth]{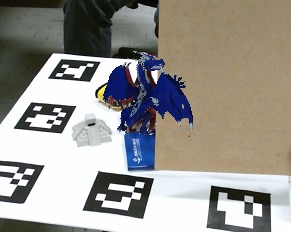} &
\includegraphics[width=.19\linewidth]{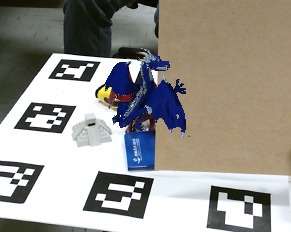} &
\includegraphics[width=.19\linewidth]{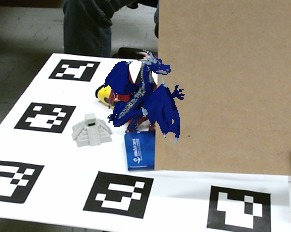} &
\includegraphics[width=.19\linewidth]{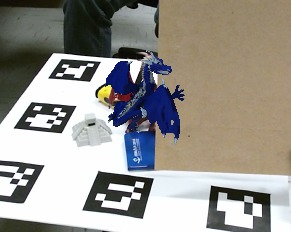} &
\includegraphics[width=.19\linewidth]{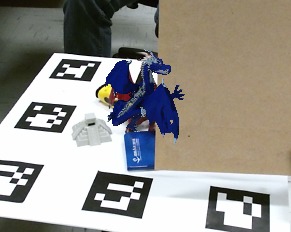} \\
\rotatebox{90}{\hspace{1em}40\% occlusion, \cite{tan-iccv-15}} & 
\includegraphics[width=.19\linewidth]{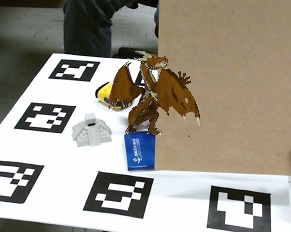} &
\includegraphics[width=.19\linewidth]{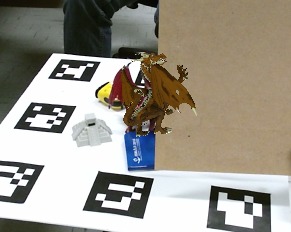} &
\includegraphics[width=.19\linewidth]{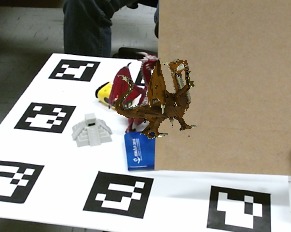} &
\includegraphics[width=.19\linewidth]{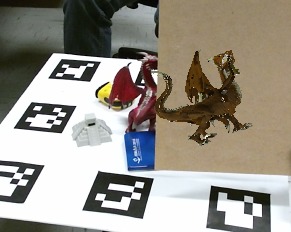} &
\includegraphics[width=.19\linewidth]{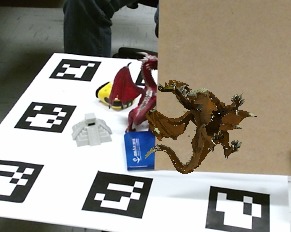} \\*[.5em]
& (a) $t = 0$ (initialization) & (b) $t = 3$ & (c) $t = 6$ & (d) $t = 9$ & (e) $t = 12$ \\
\end{tabular}
\caption{Qualitative evaluation of robustness to occlusions, comparing our approach to that of Tan et al.~\cite{tan-iccv-15}. Representative examples were chosen from the same sequences that were used to generate the plots in~\autoref{fig:occlusions-quantitative}. Each column shows the prediction immediately after initialization ($t = 0$), and $t = 3,\, 6,\, 9$, and $12$ frames later. Our method is significantly more robust to occlusion that Tan et al. See supplementary material for videos of these sequences. }
\label{fig:occlusions-qualitative}
\end{figure*}

\acknowledgments{
The authors wish to thank Aditya Shekhar for his help in generating results for~\autoref{sec:initialization}. This work was partially supported by the FRQ-NT New Researcher Grant 2016NC189939, the NSERC Discovery Grant RGPIN-2014-05314 and REPARTI Strategic Network. We gratefully acknowledge the support of Nvidia with the donation of the Tesla K40 and Titan X GPUs used for this research.}

\bibliographystyle{abbrv}
\bibliography{template}
\end{document}